\documentclass[sigconf]{acmart}
\renewcommand\footnotetextcopyrightpermission[1]{} 
\usepackage{booktabs} 
\setcopyright{rightsretained}

\acmDOI{10.475/123_4}

\acmISBN{123-4567-24-567/08/06}

\acmConference[]{Preprint}{October 2018}{USA}
\acmYear{2018}
\copyrightyear{2018}
\acmArticle{4}
\acmPrice{15.00}

\usepackage{url}  
\usepackage{graphicx}  
\usepackage{amssymb}
\usepackage[linesnumbered,algoruled,boxed,lined]{algorithm2e}
\usepackage{breakurl}
\usepackage{float}
\usepackage{amsthm}
\usepackage{thmtools, thm-restate}
\usepackage{hyperref}
\usepackage{setspace}
\usepackage{enumitem}
\usepackage{ragged2e}
\usepackage{mathtools}
\usepackage{tabu}
\usepackage{epstopdf}
\usepackage{multirow}
\usepackage{verbatim}
\usepackage{amssymb}
\usepackage{color}
\usepackage{lipsum}
\usepackage{amsfonts}
\usepackage{amsmath}
\usepackage{bm}
\usepackage{comment}
\usepackage{subfig}
\usepackage{tabularx}
\SetKwInOut{Parameter}{Parameters}

\DeclarePairedDelimiter\abs{\lvert}{\rvert}

\let\emptyset\varnothing

\newtheorem{mydefinition}{Definition}

\begin{document}

\title{Streaming Network Embedding through Local Actions}


\author{Xi Liu}
\affiliation{%
  \institution{Texas A\&M University}
}
\email{xiliu.tamu@gmail.com}

\author{Ping-Chun Hsieh}
\affiliation{%
  \institution{Texas A\&M University}
}
\email{pingchun.hsieh@tamu.edu}

\author{Nick Duffield}
\affiliation{%
  \institution{Texas A\&M University}
}
\email{duffieldng@tamu.edu}

\author{Rui Chen}
\affiliation{%
  \institution{Samsung Research America}
}
\email{rui.chen1@samsung.com}

\author{Muhe Xie}
\affiliation{%
  \institution{Samsung Research America}
}
\email{muhexie@gmail.com}

\author{Xidao Wen}
\affiliation{%
  \institution{University of Pittsburgh}
}
\email{xidao.wen@pitt.edu}
\newcommand{\nick}[1]{{\color{red}\textbf{#1}}}

%
%



\begin{abstract}

Recently, considerable research attention has been paid to network embedding, a popular approach to construct feature vectors of vertices in latent space. Due to the curse of dimensionality and sparsity in graphical datasets, this approach has become indispensable for machine learning tasks over large networks. The majority of existing literature has considered this technique under the assumption that the network is static. However, networks in many applications, including social networks, collaboration networks, and recommender systems, nodes and edges accrue to a growing network as a streaming. Moreover, high-throughput production machine learning systems require to promptly generate representations for new vertices. A small number of very recent results have address the problem of embedding for dynamic networks. However, they either rely on knowledge of vertex attributes, suffer high-time complexity, or need to be re-trained without closed-form expression. Thus the approach of adapting of the existing methods designed for static networks or dynamic networks to the streaming environment faces non-trivial technical challenges.

These challenges motivate developing new approaches to the problems of streaming network embedding. In this paper We propose a new framework that is able to generate latent features for new vertices with high efficiency and low complexity under specified iteration rounds. We formulate a constrained optimization problem for the modification of the representation resulting from a stream arrival. We show this problem has no closed-form solution and instead develop an online approximation solution. Our solution follows three steps: (1) identify vertices affected by newly arrived ones, (2) generating latent features for new vertices, and (3) updating the latent features of the most affected vertices. The generated representations are provably feasible and not far from the optimal ones in terms of expectation. Multi-class classification and clustering on five real-world networks demonstrate that our model can efficiently update vertex representations and simultaneously achieve comparable or even better performance compared with model retraining.
\end{abstract}

\maketitle

\section{Introduction}\label{01_introduction}

Recently graph representation learning, also known as graph (a.k.a network) embedding, has received considerable research attention. That is due to the fact that many real-world problems in complex systems (e.g., recommender systems, social networks, biology networks, etc.) can be modelled as machine learning tasks over large graphs. Direct representations of such graphs usually suffer from the curse of dimensionality and the sparsity problem. The idea of graph representation learning is to learn a mapping that projects each vertex in a graph to a low-dimensional and dense vector. The mapping is learned with the objective of preserving the structural information of the original graph in the geometric relationships among vertices' vector representations \cite{hamilton2017representation}. The learned graph representations are regarded as informative feature inputs to various machine learning tasks. Graph representation learning has been proven to be a useful tool for many machine learning tasks, such as vertex classification \cite{grover2016node2vec}, community detection \cite{li2018community}, link reconstruction \cite{cao2018link}, and dis-link prediction \cite{liu2018semi}.


Previous studies have proposed several prominent graph embedding methods. 
LINE \cite{tang2015line} and SDNE \cite{wang2016structural} learn graph embeddings by preserving the first- and second-order proximities in the embedded space, where the former refers to the pairwise neighborhood relationship and the latter is determined by the similarity of nodes' neighbors. The difference is that SDNE \cite{wang2016structural} uses highly non-linear functions to represent the mapping function. DeepWalk \cite{perozzi2014deepwalk} and \emph{node2vec} \cite{grover2016node2vec} capture higher-order proximities in embeddings by maximizing the conditional probability of observing the ``contextual" vertices of a vertex given the mapped point of the vertex. Here ``contextual" vertices are obtained from vertices traversed in a random walk. The crucial difference between DeepWalk and \emph{node2vec} is that \emph{node2vec} employs a biased random walk procedure to provide a trade-off between breadth-first search (BFS) and depth-first search (DFS) in a graph, which might lead to a better mapping function. \emph{struct2vec} \cite{ribeiro2017struc2vec} proposes to preserve the structural identity between nodes in the representation. To achieve this goal, it first creates a new graph based on the structural identity similarity between nodes and then applies a similar procedure to DeepWalk on the created graph. A very recent method GraphWave \cite{donnat2018embedding} leverages heat wavelet diffusion patterns to represent a node's network neighborhood.


Unfortunately, all the aforementioned studies are subject to three limitations. First, these methods have focused on representation learning for a single \emph{static} graph. However, the majority of real-world networks are naturally dynamic and continuously growing. New vertices as well as their edges form in a streaming fashion. Such networks are normally referred to as ``streaming networks" \cite{chang2016positive}. Typical examples of streaming networks include social networks \cite{bliss2014evolutionary}, academic networks and recommender systems \cite{eksombatchai2018pixie}, in which new users/scholars/customers continuously join and new friendships/coauthorships/purchases constantly happen. The above methods ignore the dynamic nature and are unable to update the vertices' representations in accordance with networks' evolution. Second, these methods are \emph{transductive}. They require that all vertices in a graph be present during training in order to generate their embeddings and thus cannot generate representations for unseen vertices. In streaming networks that constantly encounter new vertices, the \emph{inductive} capability is essential to support diverse machine learning applications. Third, the time complexity of simply retraining in these methods usually increases linearly with the number of all vertices in a network. This makes simple adaptations of the above methods through re-training in streaming networks computationally expensive, let alone the uncertainty of convergence. Indeed, the few very recent works \cite{du2018dynamic,hamilton2017inductive,liu2018semi,ma2018depthlgp} adapted from the above methods require either prior knowledge of new vertices' attributes to be inductive or require retrain on new graphs with uncertain convergence time. This also renders a challenge for many high-throughput production machine learning systems that need generating the representations of new vertices promptly. In fact, a streaming network's structure may not change substantially within a short period of time, and retraining over the entire graph is usually unnecessary.

To overcome the aforementioned limitations, we propose a novel efficient online representation learning framework for streaming graphs. In this framework, a constrained optimization model is formulated to preserve temporal smoothness and structural proximity in streaming representations. We show that the model belongs to quadratic optimization with orthogonal constraints, which in general has no closed-form solution. Therefore, we propose an online approximation algorithm that is able to inductively generate representations for newly arrived vertices and has a closed-form expression. In the online algorithm, we divide the task of streaming graph representation into three sub-tasks: identifying original vertices that are affected most by the new vertices, calculating the representations of the new vertices and adjusting the representations of the affected original vertices. Since the change of a streaming graph within a short time period, compared with the entire network, is small, the algorithm only updates the representations of a small proportion of vertices. Moreover, such an update guarantees to stop within certain time, does not require to retrain a model or wait for convergence, has low space and time complexity and thus our method is particularly suitable for high-throughput production machine learning systems. 

\noindent \textbf{Contributions}. Our research contributions are as follows:

\begin{itemize}
    \item We propose a novel online representation learning framework for streaming graphs based on a constrained optimization model. Our model simultaneously takes into consideration temporal smoothness and structural proximity. Our framework is able to calculate representations of unseen vertices without knowing their attributes.
    
    \item We devise an online approximation algorithm that is able to generate feasible representations in real time for vertices arriving in a streaming manner. This algorithm is highly efficient. In particular, it does not require retraining on the entire network or additional rounds of gradient descent. Moreover, we prove that the generated representations are still feasible in the original optimization problem and quantify their expected distance to the optimal representations as time increases.
    
    \item  We conduct extensive experiments on five real data sets to validate the effectiveness and efficiency of our model in both a supervised learning task (i.e., multi-class classification) and an unsupervised learning task (i.e., clustering). The results demonstrate that the proposed framework can achieve comparable or even better performance to those achieved by retraining \emph{node2vec}, NetMF, DeepWalk or \emph{struct2vec} with much lower running time. 
\end{itemize}

The paper is organized as follows. In Section 2, we review the existing literature of representation learning on static graphs and dynamic graphs. Meanwhile, we compare the one-step time complexity of existing dynamic graphs and our solution. We propose a formal problem formulation in Section 3 with notations provided. In Section 4, we propose our method for representation learning of graph streams. We first present the method for static graphs, and then formulate a model for dynamic graphs. We show the model has no closed-form solution, and thus finally propose an online approximation solution. In the end of the section, we quantify the performance of the approximation solution. In Section 5, we empirically evaluate the performance of the proposed solution through comparison with state-of-the-art re-train based methods in terms of both $F_{1}$ scores and running times. Finally, the concluding remarks are given in Section 6. 

\section{Related Work}\label{05_related_works}

\noindent \textbf{Static Network Embedding.} Recent developments in modeling practical problems in complex systems by machine learning tasks on large graphs have highlighted the need for graph representation learning. In graph representation learning, each vertex is mapped to a point in a low-dimensional vector space, while preserving a graph's structural information. Current studies in this direction can be roughly categorized by different types of structural information preserved in the mapping. LINE \cite{tang2015line} and SDNE \cite{wang2016structural} preserve the first- and second-order proximities, with the difference that SDNE uses highly non-linear functions to model the  mapping. Inspired by recent advances in natural language processing, DeepWalk \cite{perozzi2014deepwalk} and \emph{node2vec} \cite{grover2016node2vec} preserve higher-order proximities by maximizing the conditional probability of observing the contexts of a vertex given its representations. The crucial difference lies in that \emph{node2vec} follows a biased approach to sample contexts. \emph{struct2vec} \cite{ribeiro2017struc2vec} proposes to preserve the structural identity between nodes in the representation. To achieve this goal, it first creates a new graph based on the structural identity similarity between nodes and then follows a similar method to DeepWalk on the created graph. A very recent method GraphWave \cite{donnat2018embedding} makes use of wavelet diffusion patterns by treating the wavelets from the heat wavelet diffusion process as distributions.

\noindent \textbf{Dynamic Network Embedding.} Most of the aforementioned studies have focused on static and fixed networks. However, the majority of real-world networks evolve over time and continuously grow. New vertices as well as new edges form in a stream fashion. In view of such graphs' dynamic nature, static embedding frameworks either fail to calculate the representations of unseen vertices or need to retrain on the entire network, leading to unacceptable efficiency. There are several studies on learning representations in dynamic graphs, but very few in streaming graphs, which require higher efficiency and lower uncertainty. Liu et al. \cite{liu2018semi} and Hamilton et al. \cite{hamilton2017inductive} propose to leverage node feature information to learn an embedding function that generalizes to unseen nodes. These two methods rely on prior knowledge of new vertices' attributes and are difficult to work with only topology information. Zhou et al. \cite{zhou2018dynamic} make use of the triadic closure process, a fundamental
mechanism in the formation and evolution of networks, to capture network
dynamics and to learn representation vectors for each vertex at different time steps. Its time complexity is difficult to quantify. 

Li et al. \cite{li2017attributed} present the DANE framework consisting of both offline and online embedding modules. The offline module preserves node proximity by considering both network structure and node attributes; the online module updates embeddings with the matrix perturbation theory. Its time complexity for each step is $\mathcal{O}(k^{2}(\abs{\mathcal{V}_{t}}+k))$ (see Lemma 3.3 of the paper), where $k$ denotes the embedding dimension and $\mathcal{V}_{t}$ is the set of vertices at time $t$. Jian et al. \cite{jian2018toward} design an online embedding representation learning method based on spectral embedding, which is then used for node classification. The same limitation to the methods in \cite{li2017attributed} is its high ``one-step'' time complexity (see Theorem 2 of the paper), $\mathcal{O}\big(|\mathcal{V}_{t}|^{2}\big)$ (under the assumption that the density of the adjacency matrix is less than $1/\sqrt{|\mathcal{V}_{t}|}$). ``One-step'' here means at that time, there arrives only one vertex. Similar high time complexity can be found in \cite{zhu2018high}, where one-step complexity is $\mathcal{O}(|\mathcal{V}_{t}|k^{2}+k^{4})$ (see theorem 4.1).  Ma et al. \cite{ma2018depthlgp} takes use of Gaussian Process (GP) based non-parametric model to infer the representations for unobserved vertices. Its one-step time complexity is reduced to $\mathcal{O}(|\mathcal{V}_{t}|)$. However, it relies on re-train with uncertain waiting time for convergence; meanwhile, the complexity GP models can be very high as dimensionality increases. There are also other studies \cite{du2018dynamic} that require retraining on subgraphs containing new vertices and new edges. However, their convergence time is generally uncertain. In comparison to all above methods, we propose a solution that has closed-form expression, guarantees to output representations within certain time bounds, and has as low one-step time complexity as $\mathcal{O}(\beta)$. $\beta$ is the average degree of the graph and $\mathcal{O}(\beta)$ is achieved with $D=1$.


\section{Problem Formulation}\label{02_problem}

\begin{figure}[tbp]
\centering
\includegraphics[width=0.5\textwidth, height=.38\textwidth]{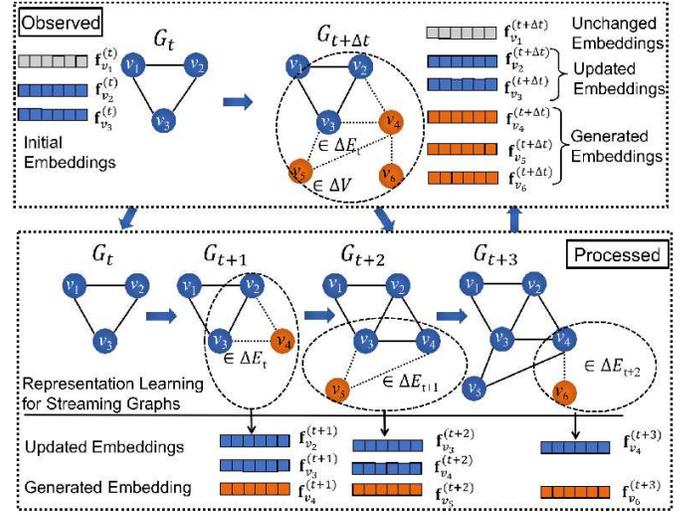}
\caption{An illustrated example of the proposed representation learning for streaming graphs with $D=1$ (only representations for the subset of neighbors are updated). Gray vectors represent unchanged representations, blue vectors represent updated representations for influenced vertices and orange vectors represent generated representations for new vertices.}
\label{fig:problem_formulation}
\end{figure}

For a streaming graph, we consider that there are new vertices and edges coming every certain time interval between time $t_0 + i\Delta t$ and time $t_0 + (i+1)\Delta t$ with $i\in \{0,1,...\}$, where $t_0$ is the initial time and $\Delta t$ is the basic time interval. In the sequel, we use  $t_i$ as the shorthand of $t_0 + i\Delta t$. The number of vertices and their edges that arrive within any $\Delta t$ can be arbitrary. Let $\mathcal{G}_{t_i} = (\mathcal{V}_{t_i}, \mathcal{E}_{t_i})$ denote the graph consisting of vertices $\mathcal{V}_{t_i}$ and edges $\mathcal{E}_{t_i}$ formed before time $t_i$. Let $\Delta \mathcal{V}_{t_{i}}$ and $\Delta \mathcal{E}_{t_{i}}$ be the vertices and their edges formed between time $t_{i}$ and $t_{i+1}$. For any time $t_i$, adding the vertices $\Delta \mathcal{V}_{t_{i}}$ and the edges $\Delta \mathcal{E}_{t_{i}}$ to the graph $\mathcal{G}_{t_i}$ leads to the new graph $\mathcal{G}_{t_{i+1}}$ at time $t_{i+1}$. For example, consider the ``Observed'' rectangle in Figure \ref{fig:problem_formulation}. Adding the vertices $v_{4}, v_{5}, v_{6}$ and their edges (depicted by dashed lines) formed between time $t_0$ and $t_1$ to $\mathcal{G}_{t_0}$ leads to $\mathcal{G}_{t_1}$. Let $\mathbf{f}^{(t_i)}_{v} \in \mathcal{R}^{k}$ be the representation of vertex $v \in \mathcal{V}_{t_i}$, where the embedding dimension $k \ll \abs{\mathcal{V}_{t_i}}$. Then at any time $t_i$, the collection of representations of vertices arrived before $t_i$ is denoted by $\{\mathbf{f}^{(t_i)}_{v}\}_{v\in \mathcal{V}_{t_i}}$. Our objective is to generate representations for the new vertices with a \emph{real-time}, \emph{low-complexity} and \emph{efficient} approach. Now we can formally define the real-time representation learning problem for streaming graphs as follows.


\begin{mydefinition} \label{def:learning} $\rm[$\emph{Real-time representation learning for streaming graphs}$\rm]$
Consider a graph $\mathcal{G}_{t_0} = (\mathcal{V}_{t_0}, \mathcal{E}_{t_0})$, which can be empty, at some initial time $t_0$. Starting from $i=0$, a collection of vertices $\Delta \mathcal{V}_{t_{i}}$ along with their edges $\Delta \mathcal{E}_{t_{i}}$ form in graph $\mathcal{G}_{t_{i}}$ between time $t_{i}$ and $t_{i+1}$ and results in a new graph $\mathcal{G}_{t_{i+1}}$ at time $t_{i}$. At any time $t_{i+1}$ with $i \in \mathbb{N}$, (1) generate representations $\{\mathbf{f}^{(t_{i+1})}_{v}\}_{v \in \Delta\mathcal{V}_{t_{i}}}$ for new vertices $\Delta\mathcal{V}_{t_i}$, and (2) update the representations $\{\mathbf{f}^{(t_{i})}_{v}\}_{v \in \mathcal{V}_{t_{i}}}$ to $\{\mathbf{f}^{(t_{i+1})}_{v}\}_{v \in \mathcal{V}_{t_i}}$ for existing vertices $\mathcal{V}_{t_{i}}$.
\end{mydefinition}

The notations used in this paper and their descriptions are listed in Table \ref{table:notations}.

\begin{table}[htbp]
    \centering
    \begin{tabular}{p{1.4cm}<{\centering}|p{5.8cm}}
    \hline
    Notations & Descriptions or Definitions
    \\ \hline
    $\mathcal{G}_{t_i}$ & Graph consisting of vertices and edges formed before time $t_i$
    \\ \hline
    $\mathcal{V}_{t_i}$ & Vertices formed before time $t_i$
    \\ \hline
    $\Delta \mathcal{V}_{t_i}$ & Vertices formed between time $t_{i}$ and $t_{i+1}$
    \\ \hline
    $\mathcal{E}_{t_i}$ & Edges formed before time $t_i$
    \\ \hline
    $\Delta \mathcal{E}_{t_i}$ & Edges formed between time $t_{i}$ and $t_{i+1}$
    \\ \hline
    $k$ & Embedding dimension
    \\ \hline
    $D$ & Depth of influence $D = \{1,2,...\}$
    \\ \hline
    $\mathbf{f}^{(t_i)}_{v}$ & $\mathcal{R}^{k}$ representation of vertex $v$ at time $t_i$
    \\ \hline
    $\mathbf{A}_{t_i}$ & Adjacency matrix of graph $\mathcal{G}_{t_i}$
    \\ \hline
    $\mathbf{D}_{t_i}$ & Diagonal matrix of graph $\mathcal{G}_{t_i}$
    \\ \hline
    $\mathbf{L}_{t_i}$ & Laplacian matrix of graph $\mathcal{G}_{t_i}$
    \\ \hline
    $\{\lambda_{j}^{(t_i)}\}_{j=1}^{\abs{\mathcal{V}_{t_i}}}$ & Eigenvalues of $\mathbf{D}_{t_i}^{-1}\mathbf{L}_{t_i}$ in ascending order $\lambda_{1}\leq \lambda_{2}\leq ... \leq \lambda_{\abs{\mathcal{V}_{t_i}}}$
    \\ \hline
    $\{\gamma_{h}^{(t)},\gamma_{s}^{(t)}\}$ & Trade-off weights between \emph{temporal smoothness} loss and \emph{graph homophily} loss at time $t$.
    \\ \hline
    $\mathbf{x}_{j}^{(t_i)}$ & $\mathcal{R}^{\abs{\mathcal{V}_{t_i}}}$ eigenvector corresponding to $\lambda_{j}$
    \\ \hline
    $\mathcal{I}_{t_i}(m)$ & Set of  vertices influenced by vertex $m$ in $\mathcal{G}_{t_i}$
    \\ \hline
    $\mathcal{N}_{t_i}(m)$ & Neighbor vertices of $m$ in $\mathcal{G}_{t_i}$
    \\ \hline
    $p^{(t+1)}_{uv}$ & Probability that $v$ influences $u$ in graph $\mathcal{G}_{t+1}$ 
    \\ \hline
    $\abs{\cdot}$ & Cardinality of a set
    \\ \hline
    $\|\cdot\|$ & The $l_{2}$ norm
    \\ \hline
    $\|\cdot\|_{F}$ & The Frobenius norm
    \\ \hline
    $tr(\cdot)$ & Trace of a matrix
    \\ \hline
    \end{tabular}
    \caption{Notations and Symbols.}
    \label{table:notations}
\end{table}

\section{Methods}\label{03_methods}
In this section, we first provide a brief introduction to representation learning for static graphs based on spectral theory and then present our model for streaming graphs. We show that our model belongs to the class of quadratic optimization problem with orthogonality constraints, which in general has no closed-form solution. We proposed an approximated solution that satisfies our wish for low-complexity, high efficiency and being real-time. The approximated solution is composed of three steps: (1) identify vertices influenced most by arrival of the new vertices, (2) generate representations of the new vertices, and (3) adjust the representations of the influenced vertices. The approximated solution is inspired by line-search method on the Stiefel manifold and influence diffusion process. We finally quantify the performance of the approximation through bounds on the expected difference between the approximated solution and the optimal solution.

\subsection{Static Graph Representation Learning}
Consider a static graph $\mathcal{G} = (\mathcal{V}, \mathcal{E})$, where $\mathcal{V} = \{v_{1},v_{2},..,v_{\abs{\mathcal{V}}}\}$. Each edge in $\mathcal{E}$ is represented by its two ends, i.e., $(v_{i},v_{j})$. The target of spectral theory based graph representation \cite{goyal2018graph} is to keep the representations of two vertices close if they are connected, a reflection of \emph{graph homophily}. Denote the adjacency matrix of $\mathcal{G}$ by $\mathbf{A}$, where $\mathbf{A}(i,j) = 1$ when $(v_{i},v_{j})\in \mathcal{E}$ and $\mathbf{A}(i,j) = 0$ otherwise. For graph $\mathcal{G}$, this target can be modelled as the optimization problem below:
\begin{align}\label{optimization}
    \min_{\mathbf{F}} \  \mathcal{L}(\mathbf{F}) &= \dfrac{1}{2}\sum_{i,j=1}^{\abs{\mathcal{V}}}\mathbf{A}(i,j)\big\|\mathbf{f}_{v_{i}} - \mathbf{f}_{v_{j}}\big\|^{2} \\ \nonumber
    s.t. \  \mathbf{F}^{\top}\mathbf{F} &= \mathbf{I}_{k\times k}\nonumber,
\end{align}
where the matrix of embeddings $\mathbf{F} \in \mathcal{R}^{\abs{\mathcal{V}}\times k}$ is:
\begin{align*}
    \mathbf{F} = \begin{bmatrix}
       (\mathbf{f}_{v_{1}})^{\top} \\
       (\mathbf{f}_{v_{2}})^{\top} \\
       \vdots \\
       (\mathbf{f}_{v_{|\mathcal{V}|}})^{\top}
     \end{bmatrix}.
\end{align*}

Denote the diagonal matrix of $\mathcal{G}$ by $\mathbf{D}$ and denote its element in the $i$-th row and $j$-th column by $\mathbf{D}(i,j)$. Then the Laplacian matrix $\mathbf{L} = \mathbf{D} - \mathbf{A}$, where $\mathbf{D}(i,i)=\sum_{j = 1}^{ \abs{\mathcal{V}}}\mathbf{A}(i,j)$ and $\mathbf{D}(i,j)=0$ for $i\neq j$. Belkin et al. \cite{belkin2002laplacian} show that Eq.(\ref{optimization}) can be solved by finding the top-$k$ eigenvectors of the following generalized eigen-problem: $\mathbf{L}\mathbf{x}=\lambda \mathbf{D}\mathbf{x}$. Let $\mathbf{x}_{1},\mathbf{x}_{2},...,\mathbf{x}_{\abs{\mathcal{V}}}$ be the eigenvectors of the corresponding eigenvalues $0=\lambda_{1}\leq \lambda_{2}\leq...\leq\lambda_{\abs{\mathcal{V}}}$. It is easy to verify that $\mathbf{1}$ is the only corresponding eigenvector for eigenvalue $\lambda_{1}$. Then the matrix of embeddings can be obtained by $\mathbf{F}=[\mathbf{x}_{2},\mathbf{x}_{3},...,\mathbf{x}_{k+1}]$. The time complexity of calculating $\mathbf{F}$ can be as high as $\mathcal{O}(k\abs{\mathcal{V}}^{2})$ without any sparsity assumption \cite{parattefast}.

\subsection{Dynamic Graph Representation Learning}

For simplicity of presentation, our explanation for dynamic graph representation learning focuses on the case where only one vertex along with its edges is added to a graph each time. In fact,
with a solution able to handle a single vertex addition at a time, the addition of multiple vertices can be solved by sequentially processing multiple single vertex additions. This is illustrated by the ``Processed'' rectangle in Figure \ref{fig:problem_formulation}. Processing the addition of $\{v_{4}, v_{5},  v_{6}\}$ in a batch can be decomposed into the sequential processing of adding $v_{4}$ at $t+1$, $v_{5}$ at $t+2$ and $v_{6}$ at $t+3$, where $t+i$ simply indicates the virtual order of processing and does not have any practical meaning. Another benefit of this decomposition is that one vertex arrival is usually regarded as a very small change to the original graph, which is consistent with the motivation of the proposed model and thus makes the approximation reasonably close to the optimal value. To be clear, in Figure \ref{fig:problem_formulation}, $\mathbf{f}_{v_{2}}^{(t+\Delta t)} = \mathbf{f}_{v_{2}}^{(t+1)}$, $\mathbf{f}_{v_{3}}^{(t+\Delta t)} = \mathbf{f}_{v_{3}}^{(t+2)}$, $\mathbf{f}_{v_{4}}^{(t+\Delta t)} = \mathbf{f}_{v_{4}}^{(t+3)}$, $\mathbf{f}_{v_{5}}^{(t+\Delta t)} = \mathbf{f}_{v_{5}}^{(t+2)}$, and $\mathbf{f}_{v_{6}}^{(t+\Delta t)} = \mathbf{f}_{v_{6}}^{(t+3)}$. Therefore, in below discussion, suppose initially at time $t_0=1$, the graph is empty and starting from $t_0=1$, there is vertex arrival between $t$ and $t+1$ for any $t\geq t_0$. Also suppose $\Delta t = 1$. Then we denote the single vertex and edges that arrive at time $t$ by $v_{t}$ and $\Delta \mathcal{E}_{t}$, respectively. Then we have,
$\mathcal{V}_{t} = \{v_{1},v_{2},...,v_{t-1}\}$ and $\mathcal{E}_{t} = \bigcup_{i=1}^{t-1} \Delta \mathcal{E}_{i}$.

To solve the problem defined in Definition~\ref{def:learning}, we propose an optimization problem that needs to be solved at time $t=2, 3, ...$. The objective function of the optimization problem is designed based on two key properties of the graph streams: \emph{temporal smoothness} and \emph{graph homophily}. First, since only one vertex and its edges arrive per time, the dynamic graph will evolve smoothly, most of the representations of the same vertices at two consecutive time steps should be close. This property is referred to as \emph{temporal smoothness}. This property has also been observed and shown to be helpful to improve representation performance in \cite{liu2018semi}. Suppose that we are at time $t+1$. Then, this property can be modelled by minimizing the following objective function at any time $t+1$:
\begin{align}\label{optimization:s}
    \mathcal{L}^{(t+1)}_{s}(\mathbf{F}_{t+1}) :&= \sum_{v_{i}\in \mathcal{V}_{t}}\big\|\mathbf{f}^{(t+1)}_{v_{i}}-\mathbf{f}^{(t)}_{v_{i}}\big\|^{2},
\end{align}
which is the summation of squared $\ell_{2}$ norm of representation difference for the same vertices in two consecutive graph snapshots $\mathcal{G}_{t}$ and $\mathcal{G}_{t+1}$. 

Second, the target of representation learning suggests that connected vertices should be embedded to close points in the latent representation space. This property is referred to as \emph{graph homophily}. This property has been reflected in the objective function and constraints of the optimization in Equation (\ref{optimization}). Thus, they should be kept for the new graph $\mathcal{G}_{t+1}$.  Formally, this property can be modelled by minimizing the following objective function at time $t+1$:
\begin{align}\label{optimization:h}
    \mathcal{L}^{(t+1)}_{h}(\mathbf{F}_{t+1}) := \dfrac{1}{2}\sum_{i,j=1}^{\abs{\mathcal{V}_{t+1}}}\mathbf{A}_{t+1}(i,j)\big\|\mathbf{f}^{(t+1)}_{v_{i}} - \mathbf{f}^{(t+1)}_{v_{j}}\big\|^{2}.
\end{align}

To take into account these two properties, we include both $\mathcal{L}^{(t+1)}_{s}$ and $\mathcal{L}^{(t+1)}_{h}$ in the final objective function and retain the constraint given in Equation (\ref{optimization}). The optimization problem to solve at time $t+1$ can be summarized as follows.
\begin{align}\label{optimization2}
    \min_{\mathbf{F}_{t+1}} \  &\mathcal{L}^{(t+1)}(\mathbf{F}_{t+1}) = \gamma^{(t+1)}_{s}\mathcal{L}_{s}^{(t+1)}(\mathbf{F}_{t+1})+\gamma^{(t+1)}_{h}\mathcal{L}_{h}^{(t+1)}(\mathbf{F}_{t+1}) \\ \nonumber
     s.t. \  &\mathbf{F}_{t+1}^{\top}\mathbf{F}_{t+1} = \mathbf{I}_{k\times k}\nonumber,
\end{align}
where, the matrix of embeddings $\mathbf{F}_{t+1} \in \mathcal{R}^{\abs{\mathcal{V}_{t+1}}\times k}$ and $\gamma^{(t+1)}_{s}$ and $\gamma^{(t+1)}_{h}$ are normalization term for the temporal smoothness loss functions $\mathcal{L}_{s}^{(t+1)}$ and graph homophily loss function $\mathcal{L}_{h}^{(t+1)}$ with

\begin{align}
    \gamma_{s}^{(t+1)} = \dfrac{1}{\abs{\mathcal{V}_{t+1}}}, \\
    \gamma_{h}^{(t+1)} = \dfrac{1}{\abs{4\mathcal{E}_{t+1}}}.
\end{align}
Consider the objective function in Equation (\ref{optimization2}). It is straightforward to observe that $\mathcal{L}_{h}^{(t+1)}(\mathbf{F}_{t+1})$ is convex in $\mathbf{F}_{t+1}$. Therefore, $\mathcal{L}^{(t+1)}(\mathbf{F}_{t+1})$ is convex if $\mathcal{L}^{(t+1)}_{s}(\mathbf{F}_{t+1})$ is also convex. This is true if we express $\mathcal{L}^{(t+1)}_{s}(\mathbf{F}_{t+1})$ by:
\begin{align}\label{smooth_loss}
    \mathcal{L}^{(t+1)}_{s}(\mathbf{F}_{t+1}) &= \sum_{v_{i}\in \mathcal{V}_{t}}\big\|\mathbf{f}^{(t+1)}_{v_{i}}-\mathbf{f}^{(t)}_{v_{i}}\big\|^{2} = \big\|\mathbf{J}_{t+1}\mathbf{F}_{t+1}-\mathbf{F}_{t}\big\|^{2}_{F} \\ \nonumber
    &= tr\Big(\big(\mathbf{J}_{t+1}\mathbf{F}_{t+1}-\mathbf{F}_{t})^{\top}(\mathbf{J}_{t+1}\mathbf{F}_{t+1}-\mathbf{F}_{t}\big)\Big).
\end{align}
where $\mathbf{J}_{t+1}\in \mathcal{R}^{\abs{\mathcal{V}_{t}}\times
\abs{\mathcal{V}_{t+1}}}$ is:
\begin{align}
    \mathbf{J}_{t+1} := \begin{bmatrix}
    1       & 0  & \dots & 0 & 0 \\
    0       & 1  & \dots & 0 & 0 \\
    \vdots & \vdots & \ddots & \vdots & \vdots \\
    0       & 0  & \dots & 1 & 0
\end{bmatrix}_{|\mathcal{V}_{t}|\times|\mathcal{V}_{t+1}|}
\end{align}

Multiplying $\mathbf{J}_{t+1}$ on the left side of $\mathbf{F}_{t+1}$ will eliminate the last row of $\mathbf{F}_{t+1}$. The expression in Equation (\ref{smooth_loss}) indicates that $\mathcal{L}^{(t+1)}_{s}(\mathbf{F}_{t+1})$ is convex in terms of $\mathbf{F}_{t+1}$. Therefore, the objective function in Equation (\ref{optimization2}) is convex over $\mathbf{F}_{t+1}$. Since the constraints in Equation (\ref{optimization2}) are orthogonality constraints, the optimization problem to solve is a general formed quadratic optimization problem under orthogonality constraints. The space defined by the orthogonal constraints is Stiefel manifold. The problem with such format has been widely studied and concluded with no closed-form solution. State-of-the-art solution is to learn the solution through Riemann gradient approach \cite{xu2018truly} or line-search method on the Stiefel manifold \cite{liu2016quadratic}, whose convergence analysis has attracted extensive research attention very recently. However, they are not suitable for streaming setting, becaus waiting for convergence brings in time uncertainty and gradient-based methods possess unsatisfied time complexity.

\subsection{Approximated Algorithm in Graph Streams}

Motivated by the aforementioned limitations, we propose an approximated solution that satisfies the low-complexity, efficiency and real-time requirement in streaming setting. The proposed approximated solution is inspired by an observation of the line-search method. The basic idea of the line-search method for the optimization problem is to search the optimal solution in the tangent space of the Stiefel manifold. We observed that line-search method based on the polar decomposition-based retraction updates the representation of a vertex through linear summation of other representations in iterations\cite{liu2016quadratic}. In our problem, that means:
\begin{align}
    \mathbf{F}^{(i+1)}_{t+1} = \Big(\mathbf{F}^{(i)}_{t+1}+\alpha_{i}\mathbf{\Gamma}^{(i)}\Big)\Big[\mathbf{I}_{k\times k}+\alpha^{2}_{i}(\mathbf{\Gamma}^{(i)})^{\top}\mathbf{\Gamma}^{(i)}\Big]^{-1/2},
\end{align}
where the superscript ``$(i+1)$'' denotes the iteration round, $\alpha_{i}$ is the step size, $\mathbf{\Gamma}^{(i)}$ is the search direction in the tangent space of the Stiefel manifold at iteration $i$, and $\mathbf{F}^{(i)}_{t+1}$ is the matrix of embedding at iteration $i$. This inspires us to generate new representation for a vertex from linear summation of original representations for other vertices. Meanwhile, the temporal smoothness in the problem indicates that the representations of most vertices would not change a lot. Therefore, to reduce the summation complexity, in the approximated solution, we propose to only update the representations of vertices that are influenced by the new vertex. We summarize the steps of the approximated solution as follows: (1) identify vertices influenced most by arrival of the new vertices, (2) generate representations of the new vertex, and (3) adjust the representations of the influenced vertices.

\begin{figure}[htbp]
\centering
\includegraphics[width=0.5\textwidth, height=.29\textwidth]{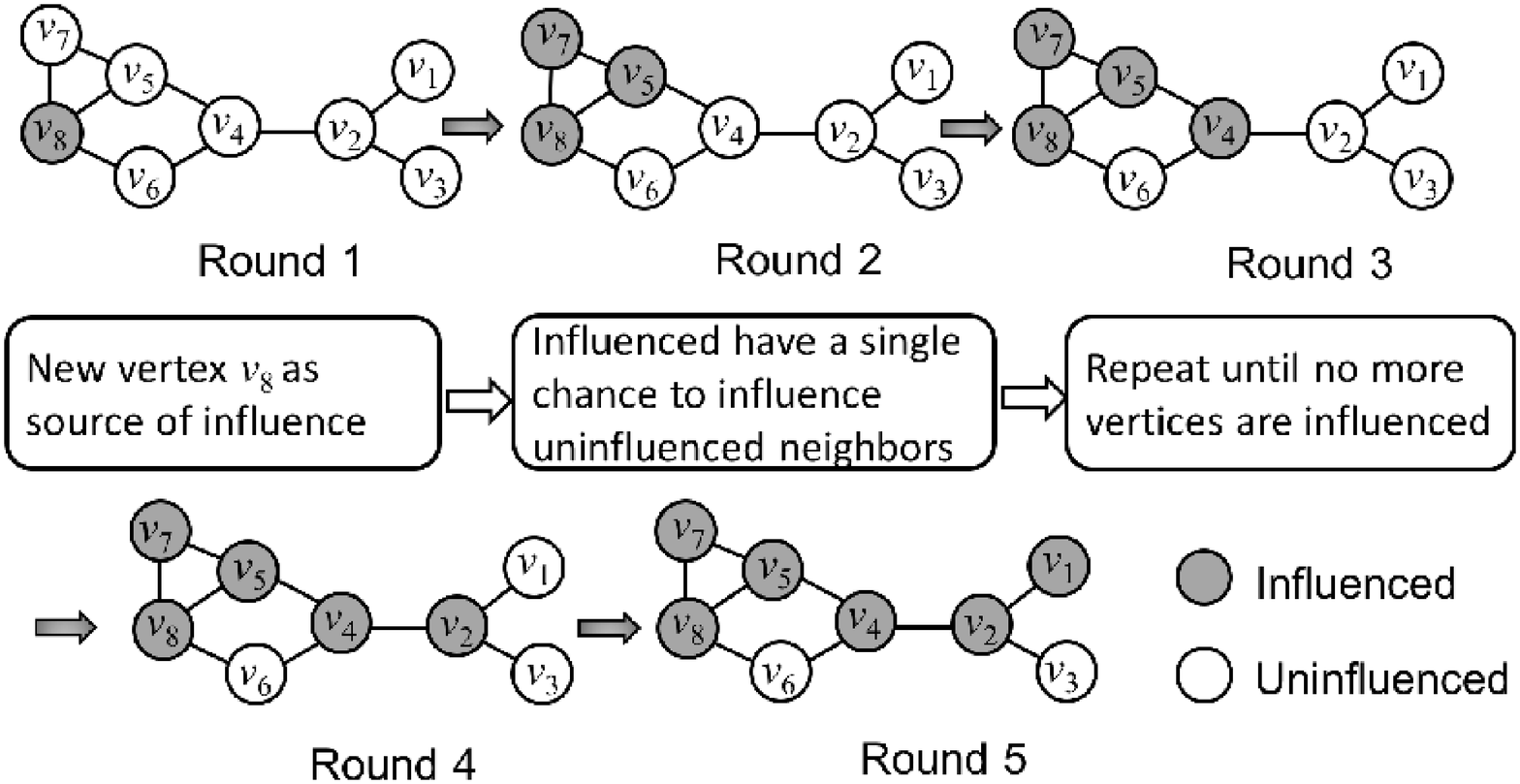}
\caption{An illustrated example of influence spread based on independent cascade model. $v_{8}$ is the new vertex. Influence depth $D = 4$.}
\label{fig:independentCascadeModel}
\end{figure}

The task for the first step can be summarized as: given a vertex, identify the set of vertices that are influenced by it. Similar problems have been widely discussed in the field ``influence propogagion'' and ``information diffusion'' (see \cite{peng2018influence} for a survey). A marriage between this field and graph representation learning has been shown very successful in a few recent works \cite{donnat2018embedding,zhang2018COSINE} for static graphs. Therefore, we apply the Weighted Independent Cascade Model (WICM), one of the most widely used models in this field, to model the influence spread. Figure \ref{fig:independentCascadeModel} provides an illustrated example of WICM, where $v_{8}$ is the new vertex and considered as the ``source'' of the influence. Suppose the influence is spread through multiple rounds, when vertex $v$ first becomes influenced at round $j$, it is given a \emph{single} chance to influence a currently uninfluenced neighbor $u$ at round $j+1$. It succeeds with a probability $p^{(t+1)}_{uv}$. The outcome is independent of the history and of $v$'s influence to other vertices. $p^{(t+1)}_{uv}$ is the probability that $v$ influences $u$ in graph $\mathcal{G}_{t+1}$ and can be estimated through
\begin{align}
    p^{(t+1)}_{uv} := \dfrac{1}{\sum_{i\in \mathcal{V}_{t+1}}\mathbf{A}_{t+1}(i,u)},
\end{align}
where the denominator is the in-degree of vertex $u$ in graph $\mathcal{G}_{t+1}$. If $u$ has multiple already-influenced neighbors other than $v$, their attempts are sequenced in an arbitrary order. The new influenced vertices will also have a single chance to influence their neighbors in next round. This process continues until no more vertices are influenced. In Figure \ref{fig:independentCascadeModel} for instance, $v_{5}$, $v_6$ and $v_7$ are possible to be influenced by $v_8$, but only $v_7$ and $v_5$ are successful. We consider that the vertices influenced most by the new vertex are all influenced ones after $D$ rounds when influence follows WICM to spread. Such set can be obtained by following Algorithm \ref{algorithm0}.

\begin{algorithm}
  \KwIn{Graph $\mathcal{G}_{t+1}$, influence depth $D$, new vertex $v_{t}$}
  \KwOut{$\mathcal{I}_{t+1}(v_{t})$ set of vertices influenced by new vertex $v_t$}
  \BlankLine
  $k = 0$, $R_{k}=\{v_t\}$;\\
  \While{ $k \leq D$}{
  $k = k +1$, $R_{k+1} = \emptyset$, $\mathcal{I}_{t+1}(v_t) = \emptyset$; \\
    \For {$v \in R_{k}$}{
        \For {$u \in \mathcal{N}_{t+1}(v)$}{
            Draw $r \sim Bernoulli(p_{uv})$;\\
            \If{$r=1$}{
                $R_{k+1}=R_{k+1}\bigcup \{u\}$;
            }
        }
    }
    {$\mathcal{I}_{t+1}(v_t) = \mathcal{I}_{t+1}(v_t) \bigcup R_{k+1}$};
  }
  \Return{ $\mathcal{I}_{t+1}(v_t)$ 
  }
\caption{Influenced vertices identification}\label{algorithm0}
\end{algorithm}

We observe two benefits to compute $\mathcal{I}_{t+1}(v_{t})$ based on Algorithm \ref{algorithm0}. First, by adjusting the value of $D$ we can control the time complexity. For instance, suppose $D=1$, Algorithm \ref{algorithm0} will stop after all neighbors of $v_t$ are visited. Second, we notice that the influence from new vertex $v_{t}$ is not equal among already-arrived vertices. It is reasonable to hope the representation of a vertex influenced less by $v_{t}$ has smaller chance to be updated than those influenced more by $v_t$. This has already been handled by WICM. As shown in Figure \ref{fig:independentCascadeModel}, compared to $v_5$, $v_1$ has smaller chance to be included in $\mathcal{I}_{t+1}(v_8)$ because to be included, all the conditions in line 7 of Algorithm \ref{algorithm0} must be true for $p_{v_{5}v_{4}}$,  $p_{v_{4}v_{2}}$ and $p_{v_{2}v_{1}}$. We also note that 
$\mathcal{I}_{t+1}(v_t)$ can be computed incrementally. Take $v_8$ as an example, storing the vertices influenced by $v_7$ and $v_{5}$ in advance, and only run the WICM from $v_8$ to $\{v_7,v_5\}$.

\begin{algorithm}
  \KwIn{Graph $\mathcal{G}_{t}$, newly arrived vertex $v_{t}$, newly arrived edges $\Delta \mathcal{E}_{t}$, matrix of embeddings  $\mathbf{F}_{t}$}
  \KwOut{Updated matrix of embeddings $\mathbf{F}_{t+1}$}
  \BlankLine
  Update graph: $\mathcal{V}_{t+1} \leftarrow \mathcal{V}_{t}\cup \{v_{t}\}$ and $\mathcal{E}_{t+1} \leftarrow \mathcal{E}_{t}\cup \Delta \mathcal{E}_{t}$;\\

  Calculate representation for new vertex $v_t$ by:\\
  $\mathbf{f}^{(t+1)}_{v_{t}} = \dfrac{1}{\abs{\mathcal{I}_{t+1}(v_{t})}}\sum_{u\in\mathcal{I}_{t+1}(v_{t})}\mathbf{f}^{(t)}_{u}$;\\
  Adjust representations for already-arrived vertices:\\
  
  \For{ $u\in\mathcal{V}_{t}$}{ 
    $\mathbf{f}^{(t+1)}_{u}=\begin{cases}
   \mathbf{f}^{(t)}_{u}-\alpha_{t+1} \mathbf{f}^{(t+1)}_{v_{t}}& u\in \mathcal{I}_{t+1}(v_{t})\\
    \mathbf{f}^{(t)}_{u}& o.w.
    \end{cases}$
  }
  \Return{ $\mathbf{F}_{t+1}$ 
  }
\caption{Representation generation and update}\label{algorithm1}
\end{algorithm}

After identification of influenced vertices, following the idea inspired by the line-search method, we generate the representation for a new vertex through a carefully designed linear summation of the influenced vertices' representations and adjust the original representations of influenced vertices. The details are illustrated in Algorithm \ref{algorithm1}. The performance of the of the algorithm will be quantified in Lemma \ref{lemma} and Lemma \ref{lemma2}. The quantity $\alpha_{t+1}$ is
\begin{align}
    \alpha_{t+1} := 1- \sqrt{1-\dfrac{1}{\abs{\mathcal{I}_{t+1}(v_{t})}}}.
\end{align}
Algorithm \ref{algorithm1} indicates that the representation of a vertex is generated when it arrives and will be updated when it is influenced by some vertices that come after it. That makes a connection between vertices that arrive at different orders and preserves the temporal pattern in later update. Since the algorithm will only update the representations of influenced vertices, different from those solutions that suffer time uncertainty from retrain, the proposed algorithm guarantees to output $\mathbf{F}_{t+1}$ after $\abs{\mathcal{I}_{t+1}(v_t)}$ rounds. Therefore, the time complexity of Algorithm \ref{algorithm1} is $\mathcal{O}\big(\abs{\mathcal{I}_{t+1}(v_t)}\big)$ and is expected to have small variance in running time. Meanwhile, the value of $\abs{\mathcal{I}_{t+1}(v_t)}$ can be controlled through changing value of $D$. That means, this algorithm provides freedom to trade-off between complexity and performance in favor of streaming setting. As discussed in Section \ref{05_related_works}, it can be as low as $\mathcal{O}(\beta)$ with $\beta$ denoting the average degree of the graph ahd $D=1$. We provide both quantification analysis for the algorithm (in Section \ref{sec:perf_quantif}) and empirical evaluation (in Section \ref{sec:sup_task} and \ref{sec:unsup_task}).

\subsection{Algorithm Quantification Analysis}\label{sec:perf_quantif}
In this subsection, we first give the lemmas, then present the proofs of the lemmas and finally remark their meanings in the contexts of representation learning for graph streams. 

\begin{lemma}\label{lemma}
Suppose at some initial time $t$, $\mathbf{F}_{t}$ is a feasible solution for the $t$ round optimization problem defined by Equation (\ref{optimization2}). Then the embedding matrix $\mathbf{F}_{t'}$ is feasible for any future $t'>t$ if each $\mathbf{F}_{t+T+1}$ is computed by inputting $\mathbf{F}_{t+T}$ to Algorithm \ref{algorithm1}, where $T=\{0,1,2,...\}$.
\end{lemma}

\begin{proof}

We only need to show that if $\mathbf{F}_{t}$ is feasible at time $t$, $\mathbf{F}_{t+1}$ returned by Algorithm \ref{algorithm1} is still feasible at time $t+1$ and this holds recursively for any $t$. Let $\mathbf{F}_{t+1}(i,j)$ be the element in $i$-th row and $j$-th column of embedding matrix $\mathbf{F}_{t+1}$ and $\mathbf{F}_{t+1}(:,j)$ be the $j$-th column of $\mathbf{F}_{t+1}$. To make $\mathbf{F}_{t+1}$ feasible, we need to prove that:
\begin{align*}
    \mathbf{F}_{t+1}(:,j)^{\top}\mathbf{F}_{t+1}(:,k) = \delta_{jk}  \ \ \forall{j,k},
\end{align*}
where
\begin{align*}
\delta_{jk}=
\begin{cases}
1& j=k\\
0& j \neq k
\end{cases}.
\end{align*}
This is sufficient to be show the following equality holds:
\begin{align*}\label{eqt:condition}
    \sum_{i=1}^{|\mathcal{I}_{t+1}(v_{t})|}&\Big(\mathbf{F}_{t+1}(i,j)-\alpha_{t} \mathbf{f}^{(t+1)}_{v_{t}}(j)\Big)\Big(\mathbf{F}_{t+1}(i,k)- \\ &\alpha_{t} \mathbf{f}^{(t+1)}_{v_{t}}(k)\Big) + \mathbf{f}^{(t+1)}_{v_{t}}(j)\mathbf{f}^{(t+1)}_{v_{t}}(k) \\
    = & \sum_{i=1}^{|\mathcal{I}_{t+1}(v_{t})|}\mathbf{F}_{t+1}(i,j)\mathbf{F}_{t+1}(i,k) \ \ \forall{j,k},
\end{align*}
where $\mathbf{f}^{(t+1)}_{v_{t}}(j)$ denotes the $j$-th element of vector $\mathbf{f}^{(t+1)}_{v_{t}}$. Now it is equivalent to prove:
\begin{align*}
    \sum_{i=1}^{\abs{\mathcal{I}_{t+1}(v_{t})}}\Big[-\alpha_{t}\Big(\mathbf{F}_{t+1}(i,j) \mathbf{f}^{(t+1)}_{v_{t}}(k)+\mathbf{F}_{t+1}(i,k)\mathbf{f}^{(t+1)}_{v_{t}}(j)\Big) + \\ \alpha_{t}^{2}\mathbf{f}^{(t+1)}_{v_{t}}(j)\mathbf{f}^{(t+1)}_{v_{t}}(k)\Big]+
    \mathbf{f}^{(t+1)}_{v_{t}}(j)\mathbf{f}^{(t+1)}_{v_{t}}(k) =  0  \ \ \forall{j,k}.
\end{align*}
Since we have:
\begin{align*}
    \sum_{i=1}^{|\mathcal{I}_{t+1}(v_{t})|}\mathbf{F}_{t+1}(i,j) = |\mathcal{I}_{t+1}(v_{t})|\mathbf{f}^{(t+1)}_{v_{t}}(j) \ \forall{j},
\end{align*}
we only need to show that:
\begin{align*}
    -2\alpha_{t} |\mathcal{I}_{t+1}(v_{t})| \mathbf{f}^{(t+1)}_{v_{t}}(j)\mathbf{f}^{(t+1)}_{v_{t}}(k) + \alpha_{t}^{2} |\mathcal{I}_{t+1}(v_{t})|\\
    \mathbf{f}^{(t+1)}_{v_{t}}(j)\mathbf{f}^{(t+1)}_{v_{t}}(k) + \mathbf{f}^{(t+1)}_{v_{t}}(j)\mathbf{f}^{(t+1)}_{v_{t}}(k) =  0  \ \ \forall{j,k}.
\end{align*}
It is easy to verify that this is true by plugging the value of $\alpha_{t}$ into the equation. Therefore, $\mathbf{F}_{t+1}$ returned by Algorithm \ref{algorithm1} is still feasible in $t+1$ round optimization problem under the orthogonal constraints. If $\mathbf{F}_{t+1}$ is used as input of Algorithm \ref{algorithm1}, then the output $\mathbf{F}_{t+2}$ will be feasible as well at $t+2$. This will recursively hold.
\end{proof}

Lemma \ref{lemma} indicates that the proposed approximated solution is a feasible solution under orthogonal constraints. Meanwhile, the proof of the feasibility is independent with $\mathcal{I}_{t+1}(v_t)$, the vertices influenced by new vertex $v_t$. That means, we are free to include as more as vertices in $\mathcal{I}_{t+1}(v_t)$ while still guaranteeing feasibility. The performance of the approximated solution is analyzed in below lemma.

\begin{lemma}\label{lemma2}
Suppose at some time $t$, for any $u\in \mathcal{V}_{t}$ there exists $\|\mathbb{E}_{\mathcal{I}_{t}}[\mathbf{f}^{(t)}_{u}-\mathbf{f}^{*}_{u}]\|^{2}<\alpha_{t}$, where expectation is taken over $\mathcal{I}_{t}$ and $\mathbf{f}^{*}_{u}$ is the optimal embedding for vertex $u$. Then the embedding $\mathbf{f^{(t^{'})}_{u}}$ satisfies $\|\mathbb{E}_{\mathcal{I}_{t'}}[\mathbf{f}^{(t')}_{u}-\mathbf{f}^{*}_{u}]\|^{2}<\alpha_{t'}$ for any future $t'>t$, if each $\mathbf{F}_{t+T+1}$ is computed by inputting $\mathbf{F}_{t+T}$ to Algorithm \ref{algorithm1}, where $T=\{0,1,2,...\}$.
\end{lemma}

\begin{proof}
Define that
\begin{align*}
    \Delta\mathcal{L}^{(t)} =  \abs{\mathcal{L}^{(t)}(\mathbf{F}_{t})-\mathcal{L}^{(t)}(\mathbf{F}_{t}^{*})},
\end{align*}
where $\mathbf{F}^{*}_{t}$ denotes the optimal embedding matrix containing $\{\mathbf{f}^{*}_{u}\}_{u\in \mathcal{V}_{t}}$. Since the expectation and summation is exchangeable, and the total loss function $\mathcal{L}^{(t)}(\cdot)$ has been normalized, we only need to show that if the conditions in the lemma hold, there exists $\mathbb{E}_{\mathcal{I}_{t+1}}[\Delta \mathcal{L}^{(t+1)}]$ $\leq 2\alpha_{t+1}$. The total loss function is consisted of two parts: the temporal smoothness loss and graph homophily loss. For the temporal smoothness loss, since only the representations of vertices in $I_{t+1}(v_t)$ are adjusted, below inequality holds for $\mathbf{F}_{t+1}$ returned by Algorithm \ref{algorithm1}: 
\begin{align*}
    \gamma_{s}^{(t+1)}\mathcal{L}_{s}^{(t+1)}(\mathbf{F}_{t+1})&= \dfrac{1}{\abs{\mathcal{V}_{t+1}}}\sum_{v_{i}\in \mathcal{I}_{t+1}(v_t)} \alpha_{t+1}\|\mathbf{f}_{v_i}^{t}\|^{2}\\
    &=\dfrac{\abs{\mathcal{I}_{t+1}(v_t)}}{\abs{\mathcal{V}_{t+1}}}\alpha_{t+1} \leq \alpha_{t+1}.
\end{align*}
Since $\mathcal{L}_{s}^{(t+1)}\geq 0$, we have $\gamma_{s}^{t+1}\Delta\mathcal{L}_{s}^{(t+1)} \leq \alpha_{t+1}$ and $\mathbb{E}_{\mathcal{I}_{t+1}}[\gamma_{s}^{t+1}\Delta\mathcal{L}_{s}^{(t+1)}]$ $\leq \alpha_{t+1}$. For the graph homophily loss, below inequality holds
\begin{align*}
    &\gamma_{h}^{(t+1)}\mathcal{L}_{h}^{(t+1)}(\mathbf{F}_{t+1}) = \dfrac{1}{8\abs{\mathcal{E}_{t+1}}}\sum_{i,j=1}^{|\mathcal{V}_{t+1}|}\mathbf{A}_{t+1}(i,j)\big\|\mathbf{f}^{(t+1)}_{v_{i}} - \mathbf{f}^{(t+1)}_{v_{j}}\big\|^{2} \\
    \leq& \dfrac{1}{8\abs{\mathcal{E}_{t+1}}}\Big(\sum_{i,j=1}^{|\mathcal{V}_{t}|}\mathbf{A}_{t}(i,j)\big\|\mathbf{f}^{(t+1)}_{v_{i}} - \mathbf{f}^{(t+1)}_{v_{j}}\big\|^{2}+\sum_{u\in\mathcal{I}_{t+1}(v_{t})} \big\|\mathbf{f}^{(t+1)}_{v_{t}} - \mathbf{f}^{(t+1)}_{u}\big\|^{2}\Big)\\
    \leq& \dfrac{1}{8\abs{\mathcal{E}_{t+1}}}\sum_{i,j=1}^{|\mathcal{V}_{t}|}\mathbf{A}_{t}(i,j)\big\|\mathbf{f}^{(t+1)}_{v_{i}} - \mathbf{f}^{(t+1)}_{v_{j}}\big\|^{2}.
\end{align*}
Taking into account the individual term of $\mathcal{L}_{h}^{(t+1)}$, we have for $v_{i}$ or $v_{j}\in\{\mathcal{I}_{t+1}(v_{t})\}$, then we have:
\begin{align*}
    &\big\|\mathbf{f}^{(t+1)}_{v_{i}} - \mathbf{f}^{(t+1)}_{v_{j}}\big\|^{2}= \big\|\mathbf{f}^{(t+1)}_{v_{i}} - \mathbf{f}^{*}_{v_{i}}\big\|^{2}+\big\|\mathbf{f}^{*}_{v_{i}} - \mathbf{f}^{*}_{v_{j}}\big\|^{2} + \big\|\mathbf{f}^{*}_{v_{j}} - \mathbf{f}^{(t+1)}_{v_{j}}\big\|^{2}  \\
    &=\big\|\mathbf{f}^{(t+1)}_{v_{i}} - \mathbf{f}^{(t)}_{v_{i}}\big\|^{2}+\big\|\mathbf{f}^{(t)}_{v_{i}} - \mathbf{f}^{*}_{v_{i}}\big\|^{2}+ \big\|\mathbf{f}^{*}_{v_{i}} - \mathbf{f}^{*}_{v_{j}}\big\|^{2}+\big\|\mathbf{f}^{(t+1)}_{v_{j}} - \mathbf{f}^{(t)}_{v_{j}}\big\|^{2}\\
    &+\big\|\mathbf{f}^{(t)}_{v_{j}} - \mathbf{f}^{*}_{v_{j}}\big\|^{2}\leq 2(\alpha_{t+1}+\alpha_{t+1}) + \big\|\mathbf{f}^{*}_{v_{i}} - \mathbf{f}^{*}_{v_{j}}\big\|^{2}.
\end{align*}
Since each edge appear twice in the summation, summing above over all edges, we have that $\gamma_{h}^{(t+1)}\Delta \mathcal{L}^{(t+1)}_{h}\leq \alpha_{t+1}$ and $\mathbb{E}_{\mathcal{I}_{t+1}}[$ $\gamma_{h}^{(t+1)}\Delta \mathcal{L}^{(t+1)}_{h}]\leq \alpha_{t+1}$. Therefore, we have $\mathbb{E}_{\mathcal{I}_{t+1}}[\Delta \mathcal{L}^{(t+1)}]$ $\leq 2\alpha_{t+1}$ and the lemma has been proved.
\end{proof}

Lemma \ref{lemma2} indicates that Algorithm \ref{algorithm1} will not amplify the expected deviation between the approximated solution and the optimal solution. We observe that $\alpha_{t+1}<1$ for any $t$, which means that the deviation is very small. Meanwhile, since the value of $\alpha_{t+1}$ depends on $\mathcal{I}_{t+1}(v_t)$, Lemma \ref{lemma2} means that through the choice of influenced vertices, Algorithm \ref{algorithm1} is able to adjust the deviation between approximated solution and optimal solution. Lemma \ref{lemma2}, in combination with Lemma \ref{lemma}, quantifies the performance of the approximated solution when the conditions in the lemmas hold.

\section{Experiments}\label{04_experiments}
In this section, we conduct experiments of both multi-class vertex classification and network clustering on five data sets to evaluate the effectiveness and efficiency of the proposed method. We use the indices of vertices as their arriving order and generate their arrived edges at each time randomly to model the streaming scenario. We evaluate our method in terms of the performance of the learning tasks and the running time to generate vertex representations. The experiments are structured to answer the following questions:
\begin{itemize}
    \item Effectiveness: compared to state-of-the-art re-train based approaches, how well the proposed approach perform in supervised learning task and unsupervised learning task under streaming setting?
    \item Efficiency: compared to state-of-the-art re-train based approaches, how faster the proposed solution is able to generate new embeddings?
    \item Scalability and stability: how stable and scalable is the proposed solution in different-scale networks?
\end{itemize}
In what follows, we first describe the data used for the experiments, then introduce the baselines which will be compared with, and finally present the results of the experiments as well as explanation and discussion.

\subsection{Data Sets}
We use the following five real data sets to validate the propose framework. All of them are publicly available
and have been widely used in previous research of both static and dynamic graph representation learning. For instance, Blog dataset have been used in \cite{perozzi2014deepwalk,grover2016node2vec,li2017attributed,ma2018depthlgp,jian2018toward,zhu2018high,qiu2018network}, Flickr dataset in \cite{perozzi2014deepwalk,tang2015line,li2017attributed,qiu2018network,jian2018toward,zhu2018high}, Cora dataset in \cite{perozzi2014deepwalk,tang2015line,grover2016node2vec}.

\begin{table}[htbp]
\centering
\caption{Dataset statistics}\label{table:statistics}
\begin{tabular}{|c|c|c|c|}
\hline
\textbf{Dataset}   & \textbf{\# of vertices}   & \textbf{\# of edges} & \textbf{\# of classes} \\\hline
Blog       & 5,196      & 171,743      & 6 \\ \hline
CiteSeer    & 3,312      & 4,732      & 6   \\ \hline
Cora    & 2,708      & 5,429      & 7   \\ \hline
Flickr    & 7,575      & 239,738      & 9   \\ \hline
Wiki    & 2,405      & 17,981      & 17   \\ \hline
\end{tabular}
\label{table:data}
\end{table}

\begin{itemize}
    \item \textbf{Blog} was collected from the BlogCatalog website, which manages bloggers and their posted blogs. Bloggers follow each other to form network edges. Bloggers categorize their blogs under some predefined classes, which are taken as the ground truth of class labels. 
    \item \textbf{CiteSeer} is a literature citation network for the selected papers indexed in CiteSeer. Papers are considered as vertices. The paper citation relations are considered as the links in the network and papers are classified into the following six classes: Agents, Artificial Intelligence, Database, Information Retrieval, Machine Learning and Human-Computer Interaction.
    \item \textbf{Cora} also represents a citation network, whose vertices represent publications from 7 classes:  Case Based, Genetic Algorithms, Neural Networks, Probabilistic Methods, Reinforcement Learning, Rule Learning, and Theory. Each link is a citation relationship between the two publications.
    \item \textbf{Flickr} was collected from Flickr, an image sharing website hosting images uploaded by users. Users in Flickr interact with others to form edges. User can subscribe different interest groups, which correspond to the class labels. The interest groups, for instance, are ``black and white photos".
    
    \item \textbf{Wiki} is a co-occurrence network of words appearing in the first million bytes of the Wikipedia dump. Each vertex in the network corresponds to a Wikipedia dump. The links between vertices are associated with hyperlinks between the two Wikipedia websites. The labels are the Part-of-Speech (POS) tags inferred by Stanford POS-Tagger.
\end{itemize}
The statistics of the datasets are summarized in Table \ref{table:statistics}.

\subsection{Compared Baseline Methods}
We compare our approach with the following four graph representation algorithms. Since they are designed for static graph representation learning, the re-train based utility of them has achieved similar performance in representation learning tasks for dynamic graphs compared to dynamic methods. Many works that handle dynamic graphs have used them as baseline methods. For instance, \emph{node2vec} in \cite{manessi2017dynamic,trivedi2018representation,zhou2018dynamic,zhu2018high}, LINE (NetMF) in \cite{li2017attributed,zhu2018high}, DeepWalk in \cite{li2017attributed,zhu2018high,zhou2018dynamic}. Since in our problem, the vertices arrive in a streaming fashion, each of the baselines is utilized through being retrained on the entire new graph. Except for those already tested in existing works, we also compare our solution with a new framework \emph{struct2vec}. For each baseline, a combination of their hyper parameters are tested and the one achieving the best performance is reported as their performance (similar with \cite{zhou2018dynamic}). To be fair, our solution use the same values for the shared hyper parameters. In the following, we refer to ``walk length" as $wl$, ``window size" as $ws$, and representation dimensions as $d$. The values of hyper parameters for baselines are obtained through grid search of different combinations: $d\in\{10, 20, ..., 200\}$, $wl \in \{10, 20, 30, 40\}$, $ws \in \{3, 5, 7, 10\}$. where the finally chosen values  are $d=90$, $wl=10$, $ws = 7$.

\begin{itemize}
    \item \textbf{NetMF} \cite{qiu2018network} obtains graph representations through explicitly factorizing the closed-form matrices 
    It has been shown to outperform LINE \cite{tang2015line} on several benchmark data sets, where LINE obtain representations through preserving the first-order and second-order proximity between vertices. 
    \item \textbf{DeepWalk} \cite{perozzi2014deepwalk} learns graph representations by preserving higher-order proximity between vertices in the embedded space. It assumes a pair
    of vertices are similar if they are close in truncated random walks. 
    \item \textbf{node2vec} is equipped with biased random walk to provide a trade-off between BFS and DFS. Compared to DeepWalk, it has a more flexible strategy to explore neighborhoods. We retrain \emph{node2vec} for new vertices with different combinations of hyper parameters: $p \in \{0.5, 1, 1.5, 2\}$, and  $q \in \{0.5, 1, 1.5, 2\}$, where $p=1$ and $q=1$ and the number of walks is $10$.
    \item \textbf{struct2vec} \cite{ribeiro2017struc2vec} learns representations by preserving the structural identity between vertices in the embedded space. It uses a hierarchical multi-layer graph to encode vertex structural similarities at different scales, and generate structural context for vertices. 
\end{itemize}

\begin{figure*}[htbp]
$\begin{array}{c c c c c}
    \multicolumn{1}{l}{\mbox{\bf }} & \multicolumn{1}{l}{\mbox{\bf }} & \multicolumn{1}{l}{\mbox{\bf }} & \multicolumn{1}{l}{\mbox{\bf }} & \multicolumn{1}{l}{\mbox{\bf }}\\
    \scalebox{0.2}{\includegraphics[width=\textwidth]{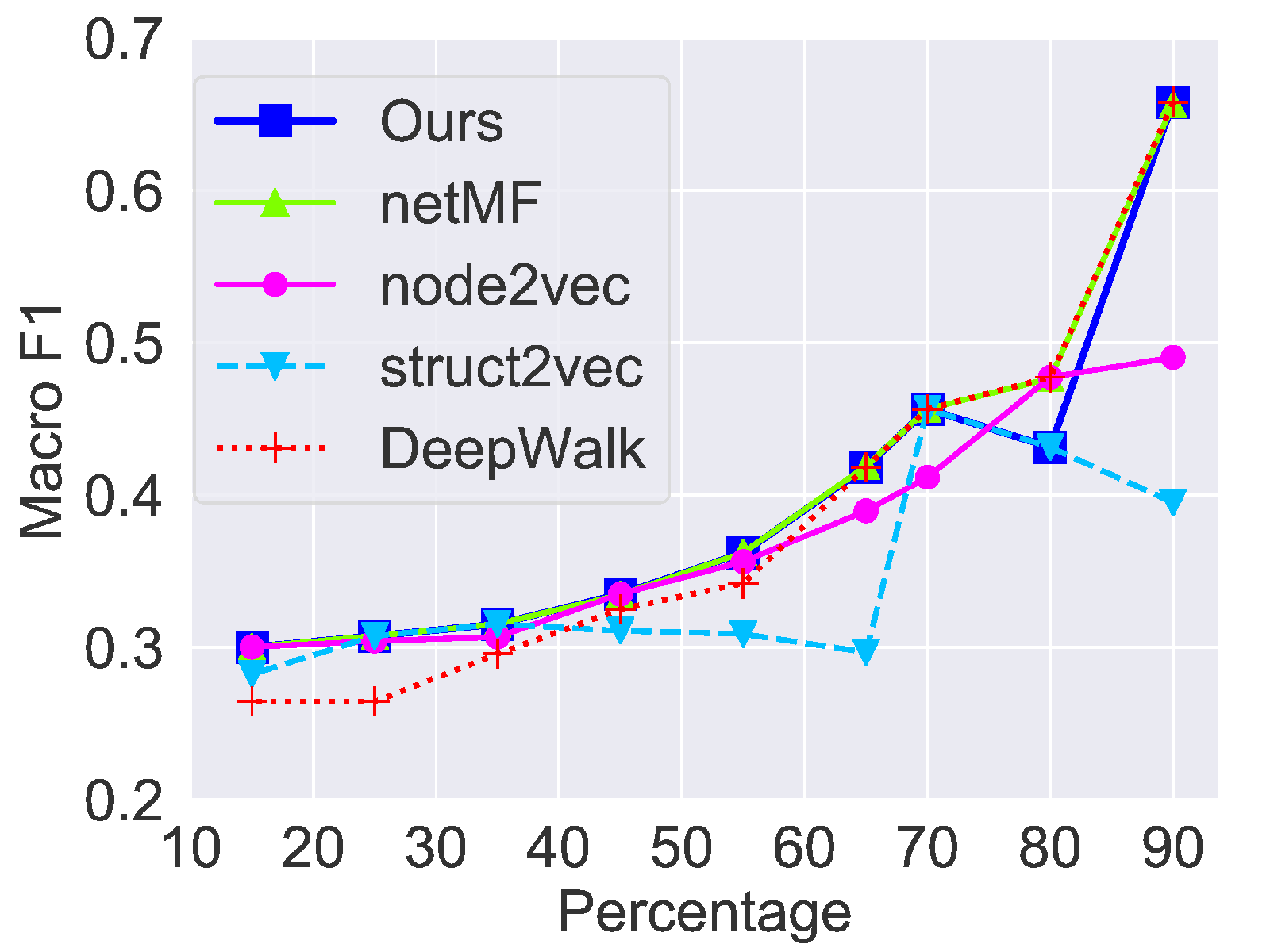}} & \hspace{-3.5mm} \scalebox{0.2}{\includegraphics[width=\textwidth]{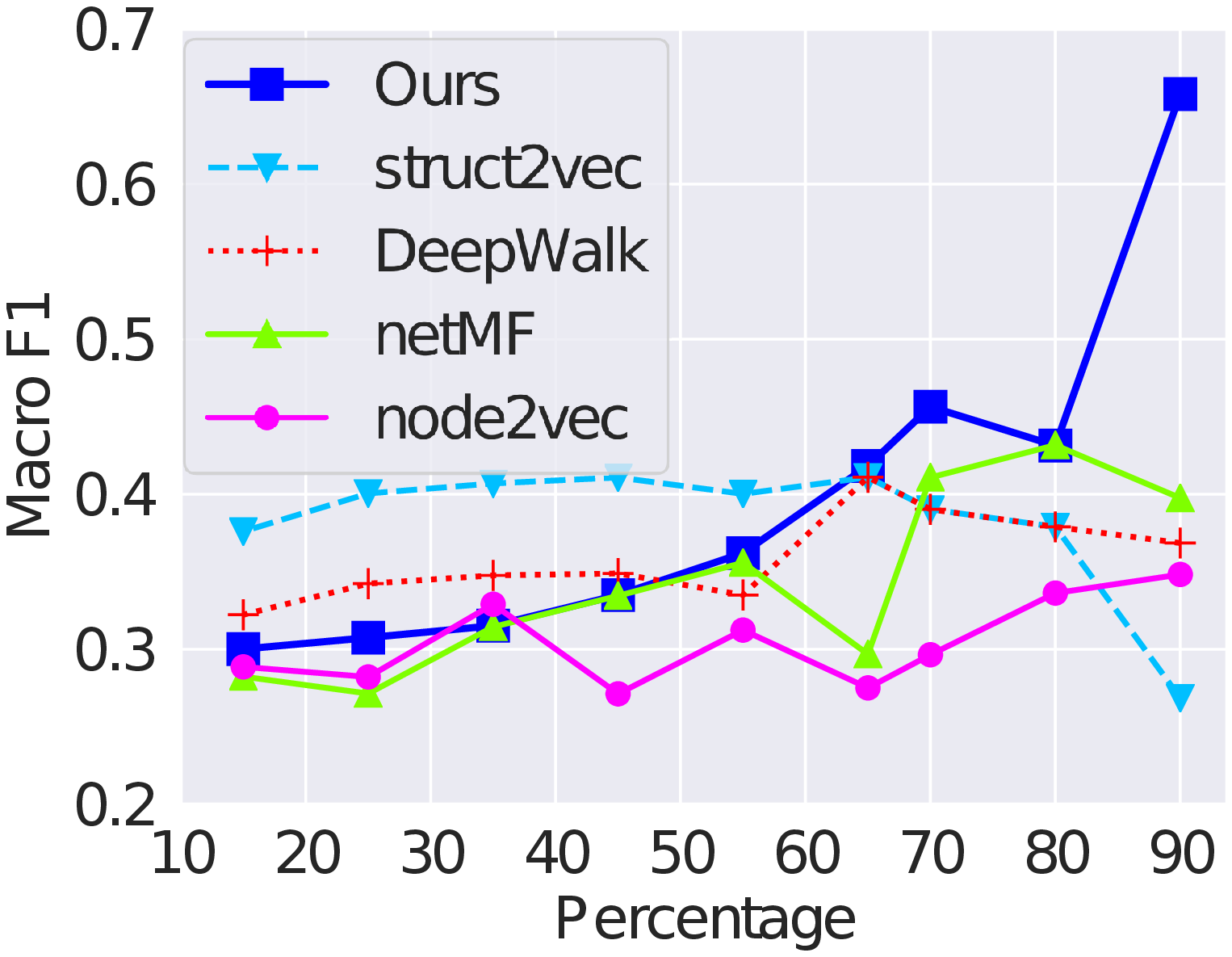}} & \hspace{-3.5mm} \scalebox{0.2}{\includegraphics[width=\textwidth]{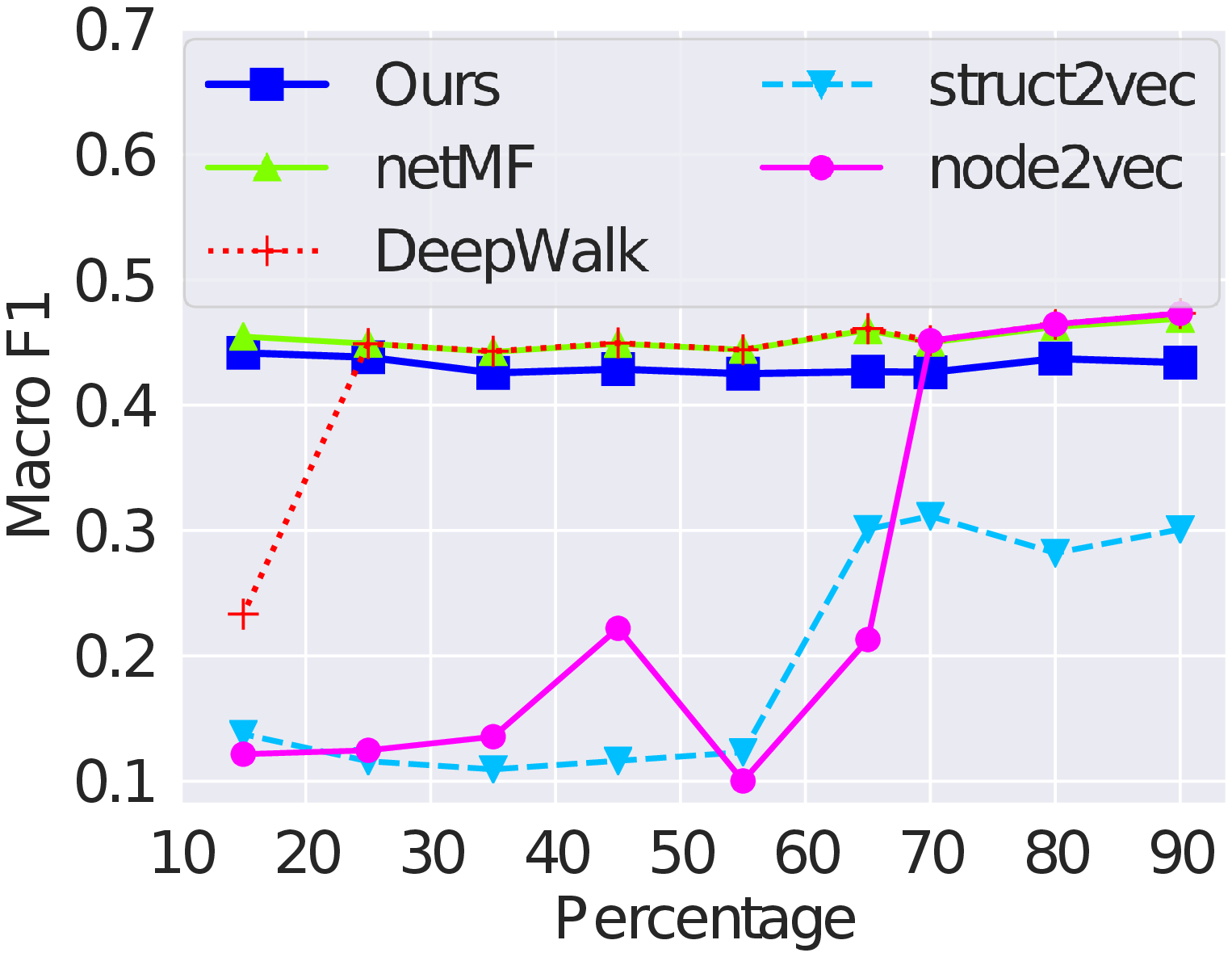}} & \hspace{-3.5mm}
    \scalebox{0.2}{\includegraphics[width=\textwidth]{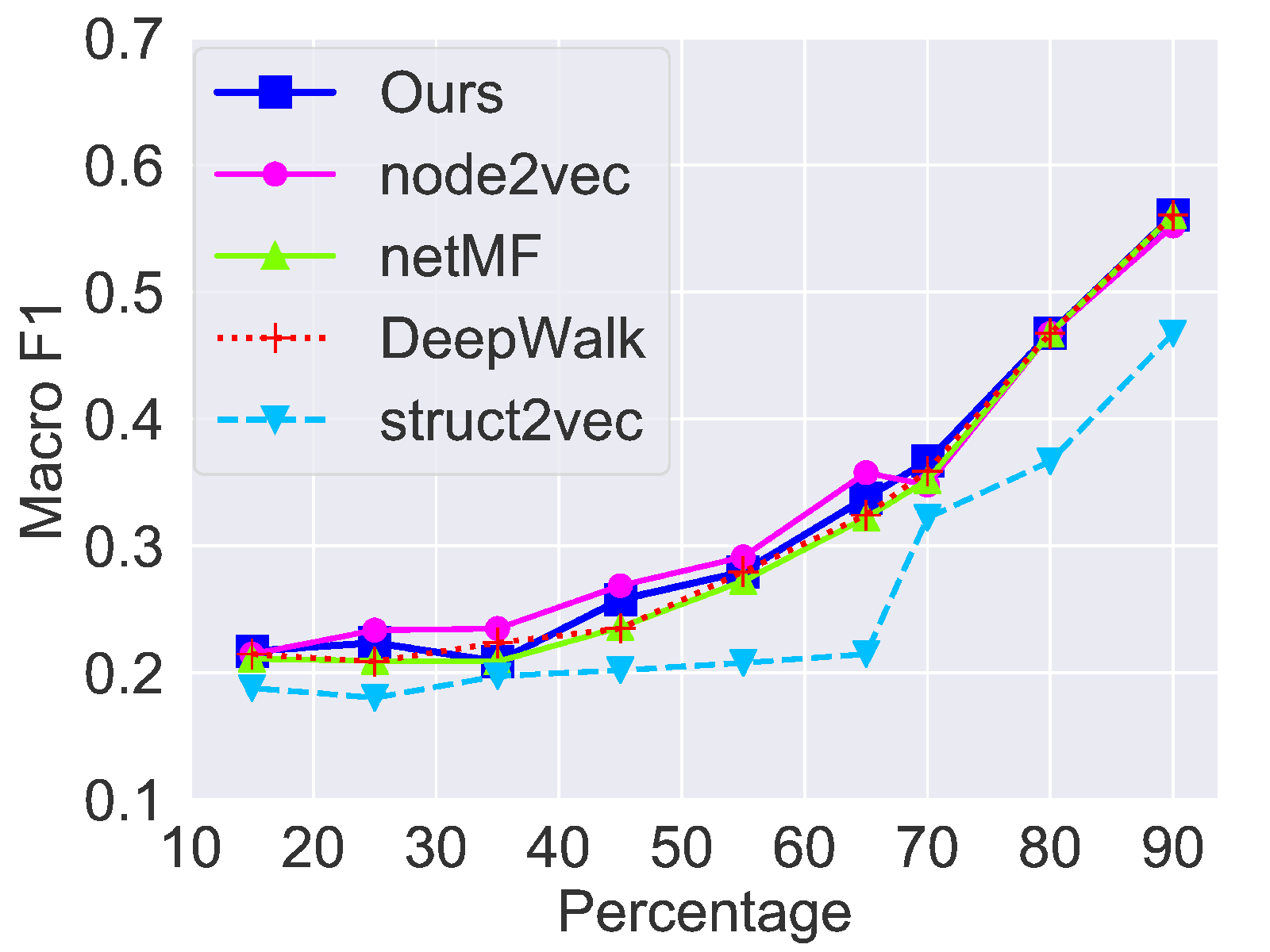}} & \hspace{-3.5mm}
    \scalebox{0.2}{\includegraphics[width=\textwidth]{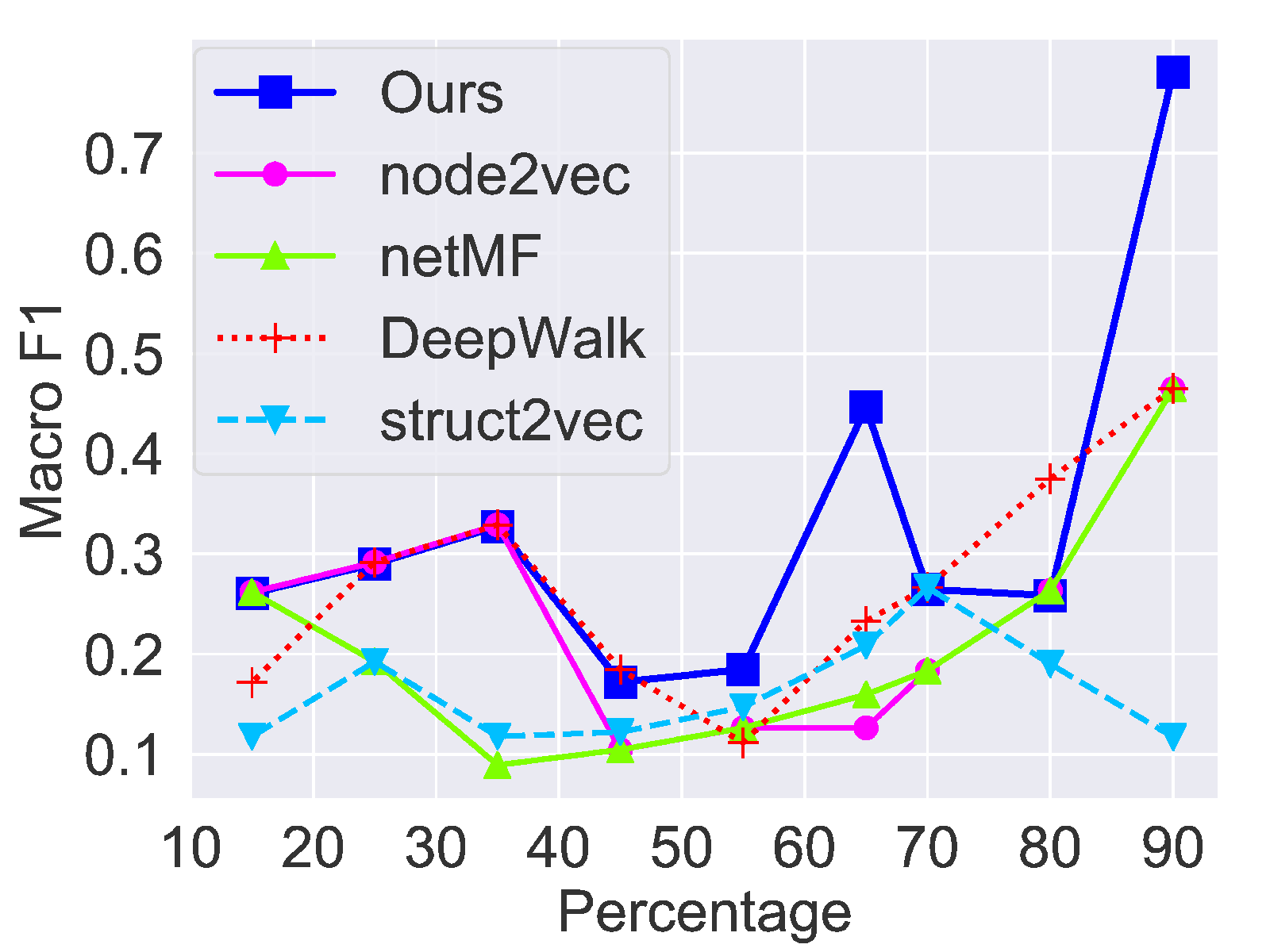}}\\ [0.01cm]
    \mbox{\small (a) Blog} & \hspace{-2mm} \mbox{\small (b) CiteSeer} & \hspace{-2mm} \mbox{\small (c) Cora} & \hspace{-2mm} \mbox{\small (d) Flickr} & \hspace{-2mm} \mbox{\small (e) Wiki}\\[-0.2cm]
    \end{array}$
\caption{Comparison of vertex multi-class classification performance in Macro-$F_{1}$ with $D=1$.}
\label{fig:macro}
\end{figure*}

\begin{figure*}[htbp]
$\begin{array}{c c c c c}
    \multicolumn{1}{l}{\mbox{\bf }} & \multicolumn{1}{l}{\mbox{\bf }} & \multicolumn{1}{l}{\mbox{\bf }} & \multicolumn{1}{l}{\mbox{\bf }} & \multicolumn{1}{l}{\mbox{\bf }}\\
    \scalebox{0.2}{\includegraphics[width=\textwidth]{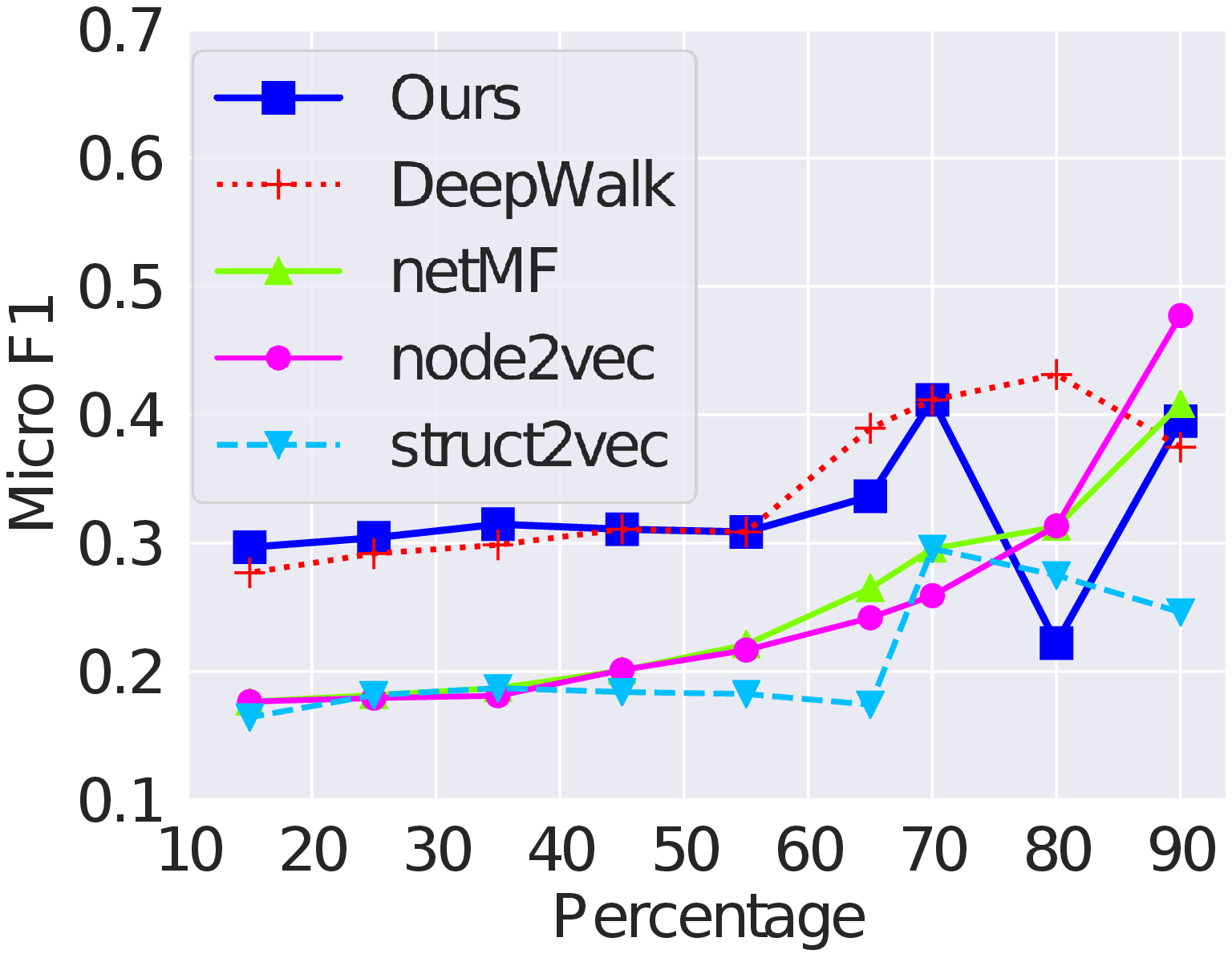}} & \hspace{-3.5mm} \scalebox{0.2}{\includegraphics[width=\textwidth]{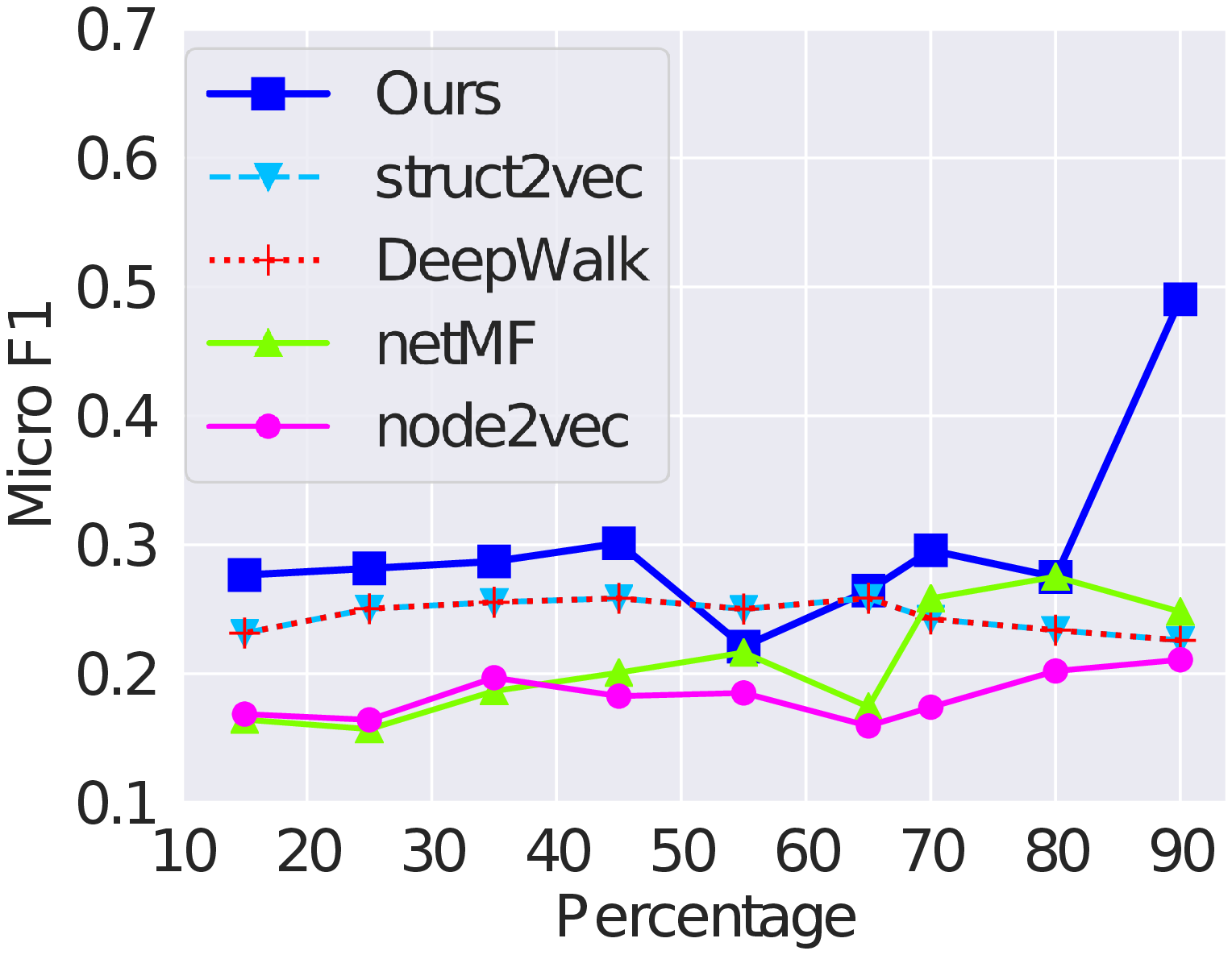}} & \hspace{-3.5mm} \scalebox{0.2}{\includegraphics[width=\textwidth]{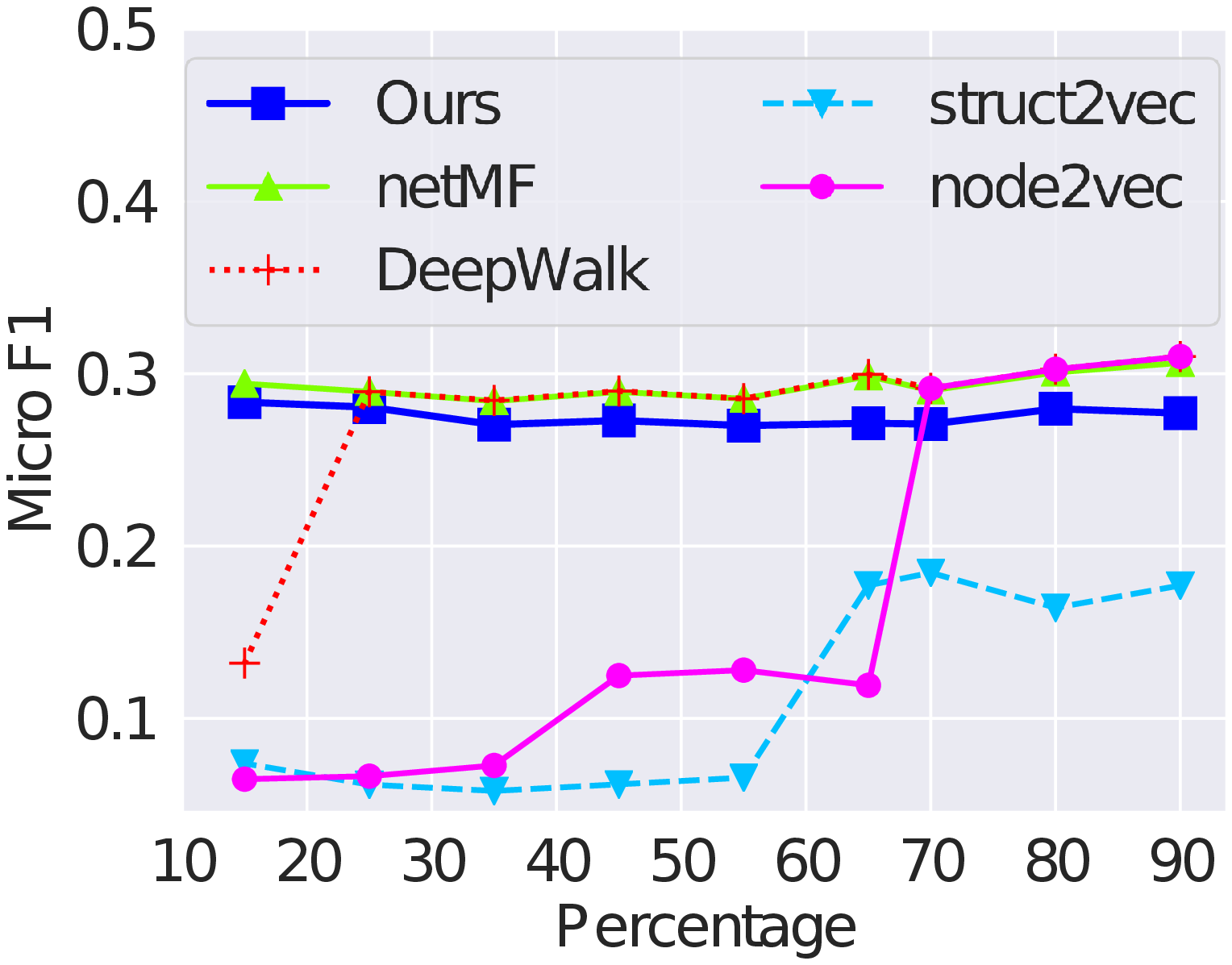}} & \hspace{-3.5mm}
    \scalebox{0.2}{\includegraphics[width=\textwidth]{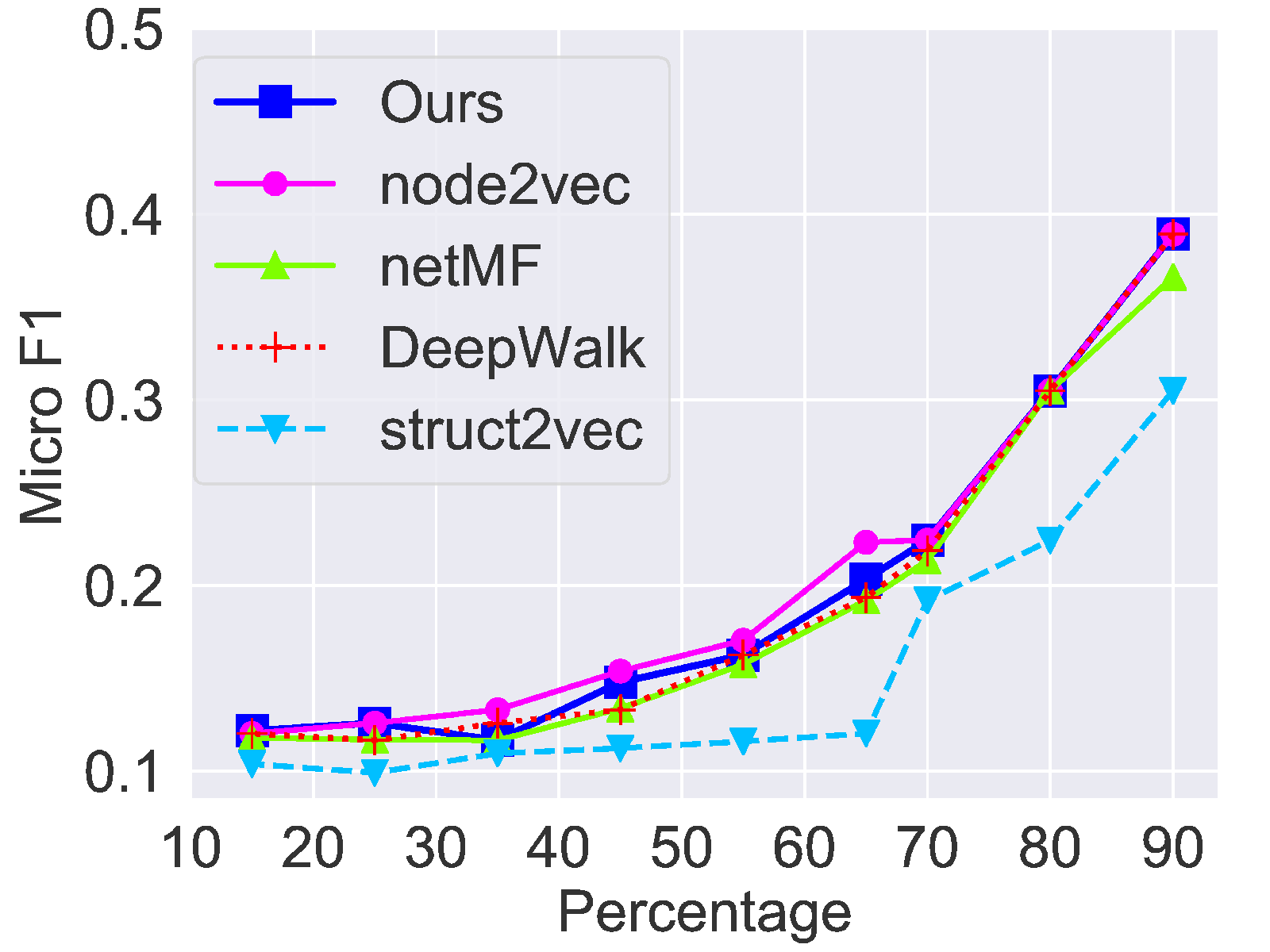}} & \hspace{-3.5mm}
    \scalebox{0.2}{\includegraphics[width=\textwidth]{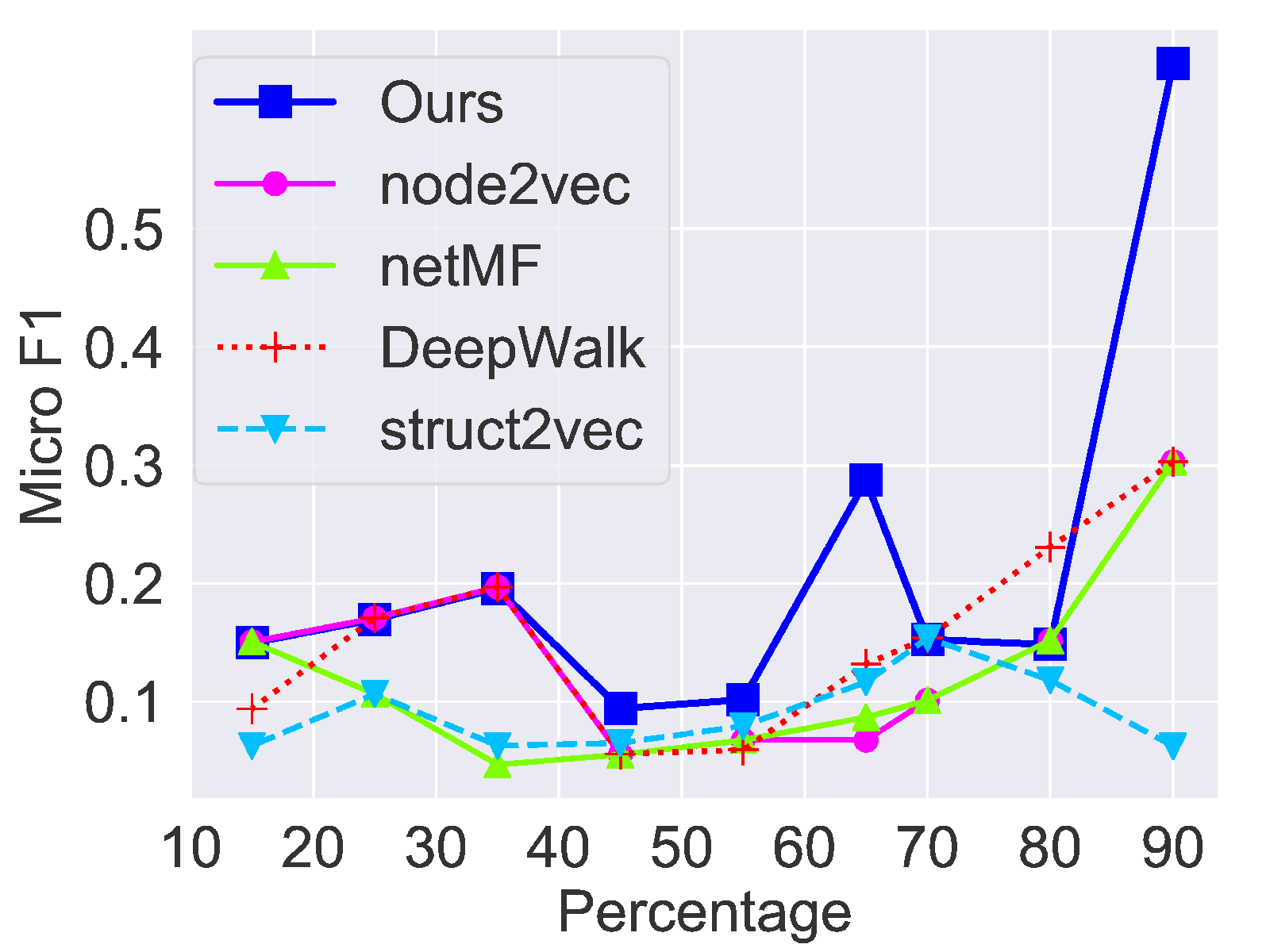}}\\ [0.01cm]
    \mbox{\small (a) Blog} & \hspace{-2mm} \mbox{\small (b) CiteSeer} & \hspace{-2mm} \mbox{\small (c) Cora} & \hspace{-2mm} \mbox{\small (d) Flickr} & \hspace{-2mm} \mbox{\small (e) Wiki}\\[-0.3cm]
    \end{array}$
\caption{Comparison of vertex multi-class classification performance in Micro-$F_{1}$ with $D=1$.}
\label{fig:micro}
\end{figure*}
\subsection{Supervised Tasks - Vertex Classification}\label{sec:sup_task}
To evaluate the effectiveness and efficiency of the proposed model, we first compare the performance of our solution with different baseline methods on the vertex classification task. The vertex representations are fed into a one-vs-rest logistic regression classifier with L2 regularization. Since graph representation learning is unsupervised and we evaluate it on a supervised learning task, we clarify the meaning of the percentage used for training here. Suppose we use $20\%$ for training, it means that for the first $20\%$ arrived vertices, we use the offline spectral theory based method in Equation (\ref{optimization}) to generate the initial representations and then follow Algorithm \ref{algorithm1} to learn representations for vertices arriving thereafter. Then the classifier for the proposed solution will be trained on the representations for the first $20\%$ arrived vertices and tested on the remaining $80\%$ arrived vertices. For a vertex that arrive after the training phase, only its representation obtained upon arrival is used in testing, even later this might be updated.  In comparison, baseline methods are allowed to retrain on vertices that arrive later. To be fair, the classifiers for the baselines and the proposed solution are required to be trained and tested on the same data set. 

\subsubsection{Effectiveness Discussion}\label{effectiveness}

We use Micro-$F_{1}$ and Macro-$F_{1}$ as the evaluation metrics. Figure \ref{fig:macro} and Figure \ref{fig:micro} compare the Macro-$F_{1}$ and Micro-$F_{1}$ performance, respectively, with varying percentages of data used for training. We observe that overall, the performance of both the proposed solution and the baselines improves as the percentage of training increases. This is easy to understand from the perspective of Lemma \ref{lemma2}, as the larger percentage is used for training, the more probable that the conditions in Lemma \ref{lemma2} are satisfied. It can be also observed that our solution achieves almost the same or even slightly better Micro-$F1$ and Macro-$F_{1}$ under varying percentages. For example, on CiteSeer, Cora and Wiki data set, our method outperforms almost all baseline methods in terms of Micro-$F_{1}$ scores for most percentages. On Blog and Flickr data set, our method achieves almost the same Micro-$F_{1}$ and Macro-$F_{1}$ for most percentages. 

The reasons that re-train based baselines sometimes are slightly worse than our method are two-fold: (1) our method captures the temporal drift of the representations through adding $\mathcal{L}^{(t)}_{s}$ into the loss function, while existing static methods ignore that information; (2) baseline methods rely on random sampling of contextual vertices to learn the representation. However, since when a vertex arrives, its neighbors can be very sparse, it might be difficult for random sampler to extract enough contextual vertices for them. Instead, the proposed method is enforced to involve influenced vertices through taking use of weighted independent cascade model. Moreover, the baselines depend on contextual vertices; in streaming setting, however only the already-arrived vertices can be selected as contexts; vertices that arrive later cannot be utilized to improve learning results. All those effects slightly degrade the performance of baseline methods at some percentages.

\begin{figure*}[t]
$\begin{array}{c c c c c}
    \multicolumn{1}{l}{\mbox{\bf }} & \multicolumn{1}{l}{\mbox{\bf }} & \multicolumn{1}{l}{\mbox{\bf }} & \multicolumn{1}{l}{\mbox{\bf }} & \multicolumn{1}{l}{\mbox{\bf }}\\
    \scalebox{0.2}{\includegraphics[width=\textwidth]{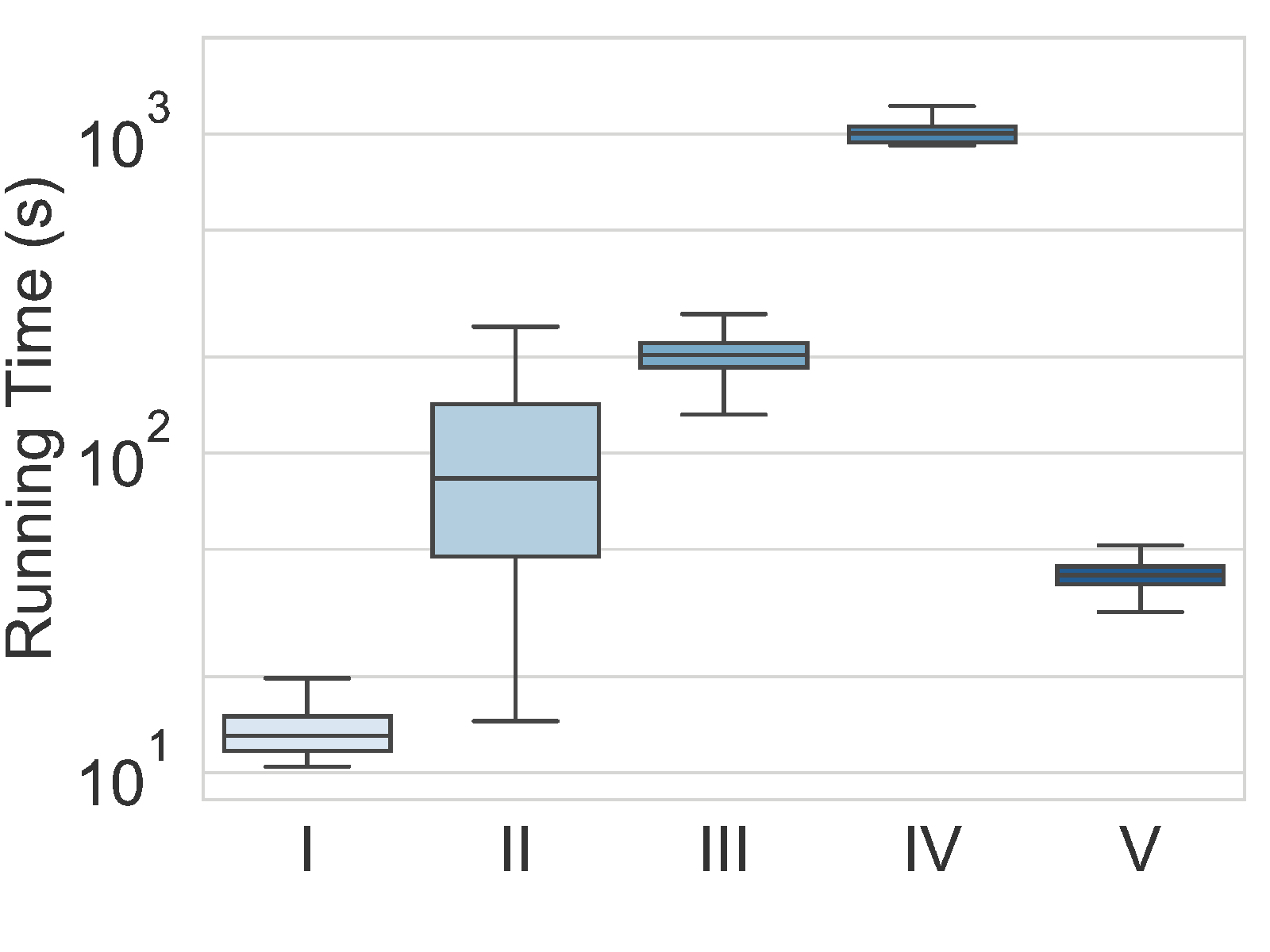}} & \hspace{-3.5mm} \scalebox{0.2}{\includegraphics[width=\textwidth]{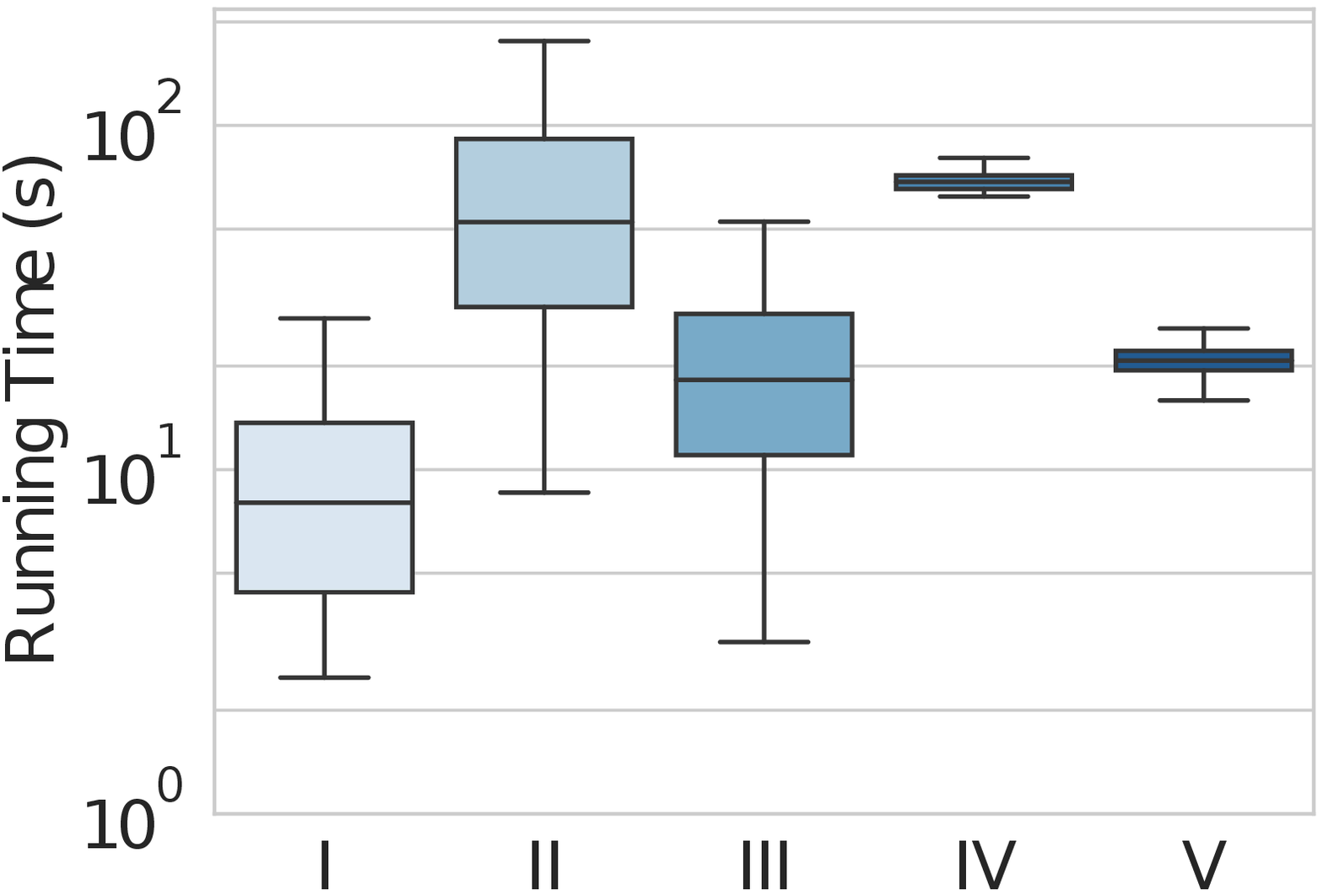}} & \hspace{-3.5mm} \scalebox{0.2}{\includegraphics[width=\textwidth]{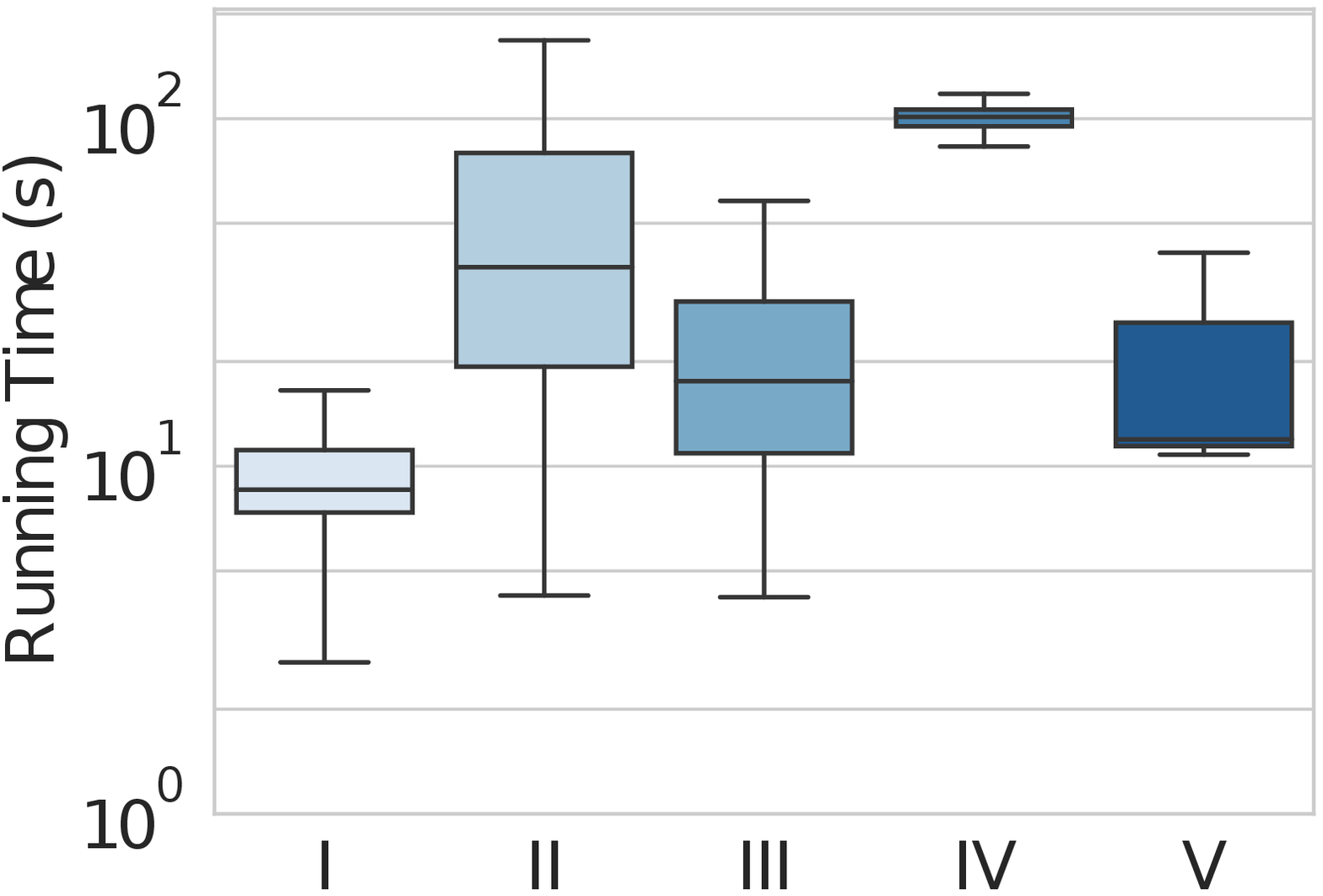}} & \hspace{-3.5mm}
    \scalebox{0.2}{\includegraphics[width=\textwidth]{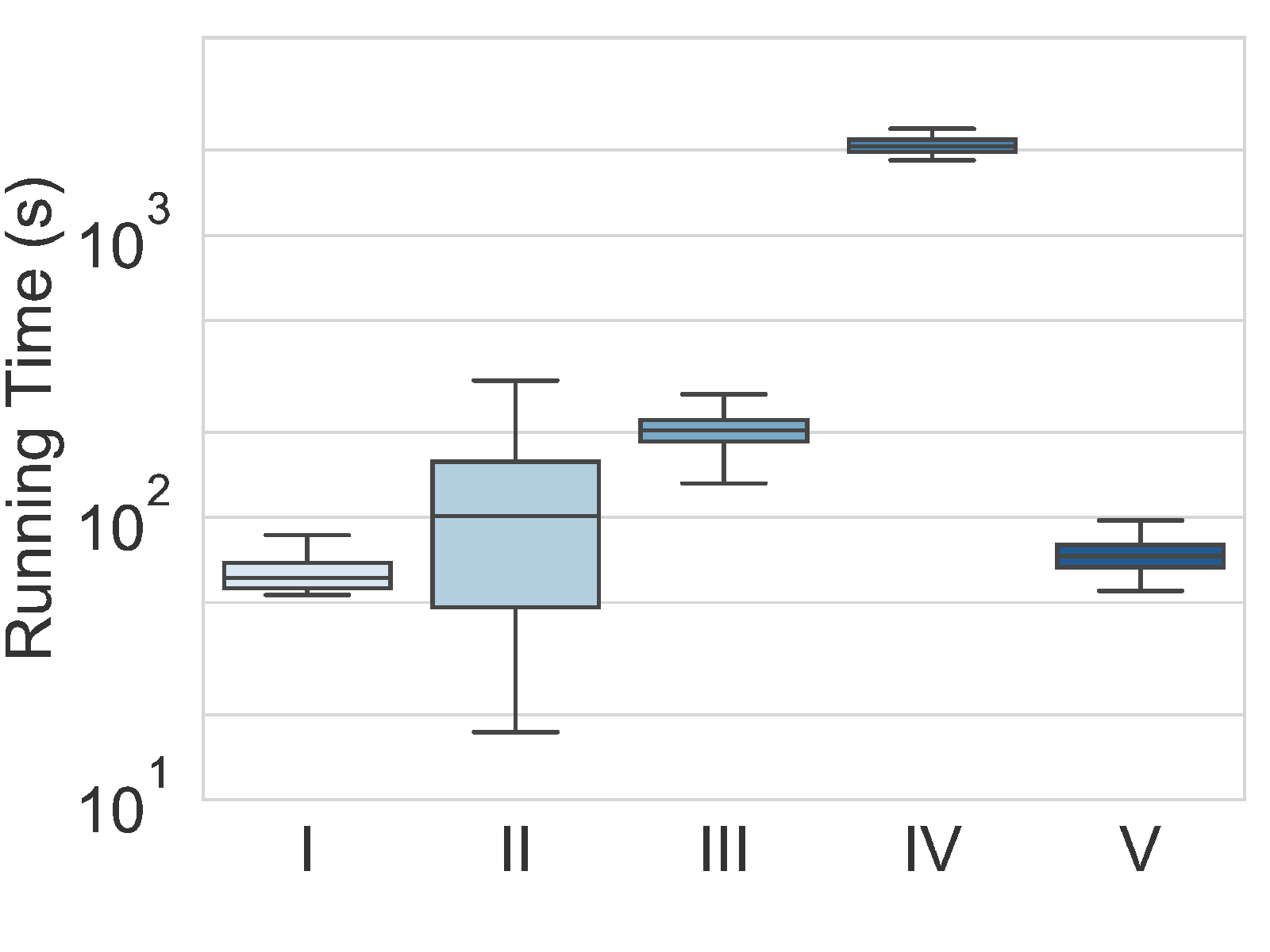}} & \hspace{-3.5mm}
    \scalebox{0.2}{\includegraphics[width=\textwidth]{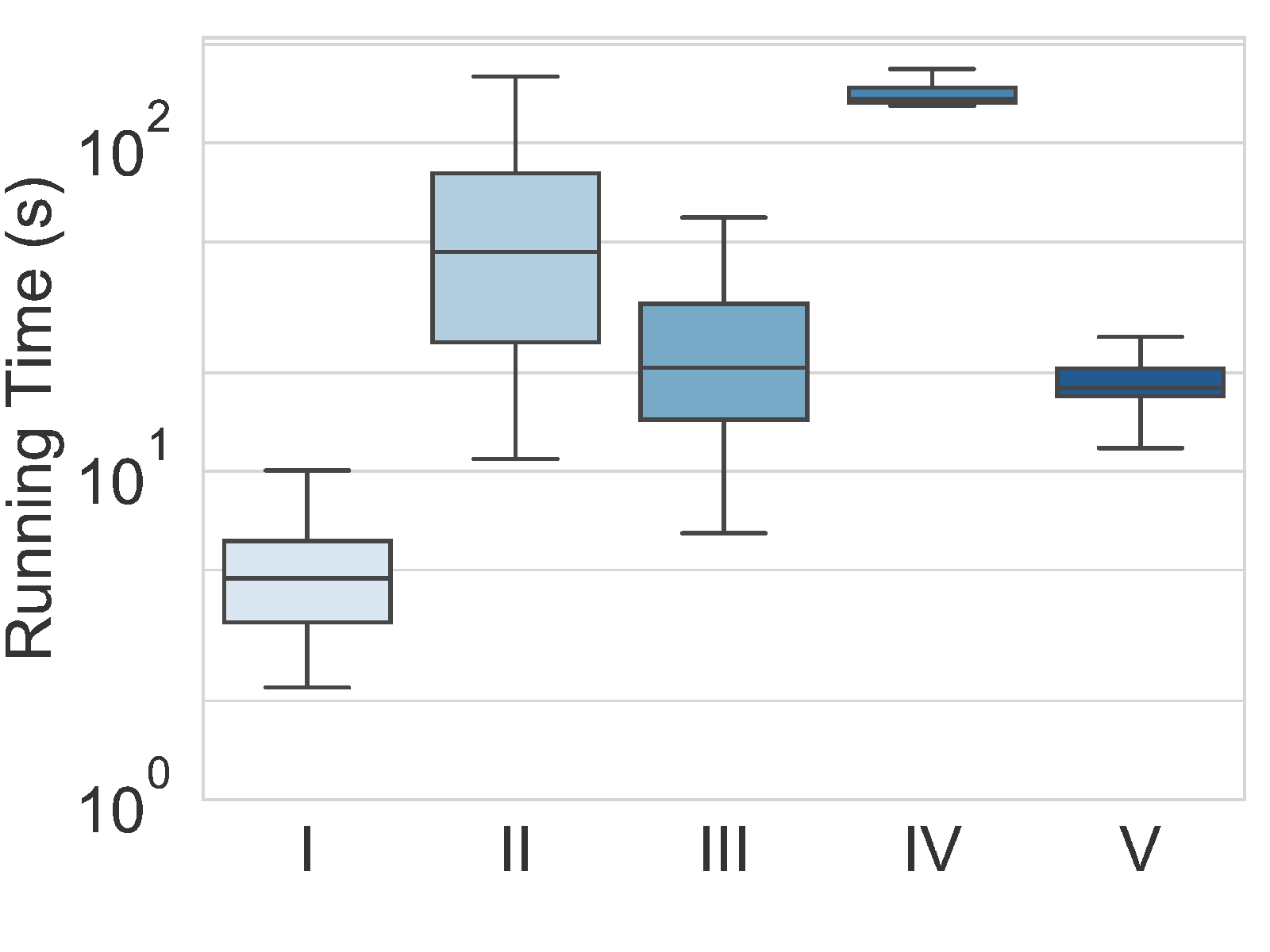}}\\ [0.0cm]
    \mbox{\small (a) Blog} & \hspace{-2mm} \mbox{\small (b) CiteSeer} & \hspace{-2mm} \mbox{\small (c) Cora} & \hspace{-2mm} \mbox{\small (d) Flickr} & \hspace{-2mm} \mbox{\small (e) Wiki}\\[-0.2cm]
    \end{array}$
\caption{Comparison of running time in seconds (I-Ours, II-DeepWalk, III-Node2Vec, IV-Struct2Vec, V-NetMF).}
\label{fig:running_time}
\end{figure*}

\begin{table*}[h]
\centering
\caption{Comparison of performance on clustering \%}
{\small
\begin{tabular}{|l|c|c|c|c|c|c|c|c|c|c|}
\hline
           & \multicolumn{2}{c|}{Blog}                            & \multicolumn{2}{c|}{CiteSeer}                               & \multicolumn{2}{c|}{Wiki}                                   & \multicolumn{2}{c|}{Cora}                                   & \multicolumn{2}{c|}{Flickr}                                 \\\hline
           & \multicolumn{1}{l|}{Completeness}
           & \multicolumn{1}{l|}{NMI} &  \multicolumn{1}{l|}{Completeness} &
           \multicolumn{1}{l|}{NMI} &
           \multicolumn{1}{l|}{Completeness} & 
           \multicolumn{1}{l|}{NMI} &
           \multicolumn{1}{l|}{Completeness} &
           \multicolumn{1}{l|}{NMI} & \multicolumn{1}{l|}{Completeness} &
           \multicolumn{1}{l|}{NMI} \\\hline
Ours       & 16.48                            & \textbf{26.71}          & 16.48                           & 22.46                   & 16.48                            & 26.71                   & \textbf{34.50}                   & \textbf{34.62}          & \textbf{16.53}                   & \textbf{20.44}          \\\hline
DeepWalk   & 17.77                            & 20.01                   & 17.77                            & 20.01                   & 11.78                            & 11.72                   & 34.62                            & 34.30                   & 16.46                            & 17.31                   \\\hline
netMF      & 0.55                             & 0.67                    & 0.25                             & 0.27                    & 6.85                             & 7.17                    & 6.99                             & 7.49                    & 1.91                             & 2.05                    \\\hline
node2vec   & 16.78                            & 22.93                   & 17.55                            & 22.93                   & 16.60                            & 27.77                   & 31.77                            & 31.60                   & 15.82                            & 21.65                   \\\hline
struct2vec & 4.35                             & 6.44                    & 2.34                             & 2.44                    & 3.57                             & 4.74                    & 9.80                             & 7.94                    & 7.23                             & 8.04                   \\\hline
\end{tabular}
}
\label{tab:clustering}
\end{table*}

\subsubsection{Efficiency Discussion}
We empirically evaluate the running times of different methods in Figure \ref{fig:running_time}. Note that the y-axis is in log scale. The running times for the proposed method count both the times to generate training representations and testing representations. Since the variation of training percentages influence the performance, we collect running times over all training percentages. The running times for the baseline methods count the times to generate representations for the final graph. We observe that, in general, the running time of our solution is much smaller, let alone if we compare them under one-step generation, the running time of our solution will be further shorter. Meanwhile, since the variance of running times is over different training percentages and comparatively small in most figures, the proposed method enjoys reasonably good stability to the training percentages. Moreover, along with the statistics in Table \ref{table:statistics}, we note the increased running times of the proposed solution is not as large as other baseline methods when the network size increases. That indicates, the scalability of the proposed solution is empirically good. The advantages of the proposed solution is because to generate the representations for newly arrived vertices, all baseline methods need to retrain over all arrived vertices and wait for convergence, while our solution only need to update over its neighbors and guarantee to stop after certain steps.

\subsection{Unsupervised Tasks - Network Clustering}\label{sec:unsup_task}
Next, we assess the effectiveness of different vertex representations on an unsupervised learning task - network clustering. Since the variation of training percentages influence the performance, we compare the average clustering performance over all training percentages. We use the same representations used in vertex classification task. Thus our method's running time is also illustrated as shown in Figure \ref{fig:running_time}. We perform $K$-means clustering based on the representations generated by our method and different baselines. $K$ is chosen to equal the number of classes in each data set.  $K$-means algorithm is repeated $10$ times and the average results are reported since $K$-means may converge to the local minima due to different initialization. We use normalized mutual information (NMI) and completeness score as the performance metrics. They help quantify how close the clustering results are to the ground-truth class belongings, e.g., whether most same-class vertices are assigned to the same cluster. The computation of the two evaluation metrics can be expressed below:
\begin{align*}
    NMI :&= \dfrac{2I(C;K)}{H(C)+H(K)}, \\
    Completeness :&= 1-\dfrac{H(K|C)}{H(K)},
\end{align*}
where $C$ denotes the class assignment, $K$ denotes the clustering assignment, $I(\cdot,\cdot)$ is the mutual information and $H(\cdot)$ is the entropy. The results are summarized in Table \ref{tab:clustering}. Again it can be seen that our method achieves comparable or even better performance. For example, our method achieves the best performance on Cora and Flickr. As a remind, Flickr network is the largest among all. Please refer Section \ref{effectiveness} for discussion of the reasons.

\section{Concluding Remarks}\label{06_conclusions}
We proposed an efficient online representation learning framework for graph streams, in which, new vertices as well as their edges form in a streaming fashion. The framework is inspired by incrementally approximating the solution to a constructed constrained optimization problem, which preserves temporal smoothness and structural proximity in resultant representations. Our solution has a closed form, high efficiency, and low complexity. We proved that the approximated solution is still feasible and is guaranteed to be close to the optimal representations in terms of the expectation. Meanwhile, the upper bound of the deviation can be controlled by a hyper parameter in our solution. To validate the effectiveness of our model and learning algorithm, we conducted experiments on five real-world networks for both supervised and unsupervised learning tasks (multi-class classification and clustering) with four baseline methods. Experimental results demonstrate that compared with several state-of-the-art techniques, our approach achieves comparable performance to that of retraining the entire graph with substantially less running time.

\clearpage
\newpage
\bibliography{ref}

\bibliographystyle{ACM-Reference-Format}
\end{document}